\documentclass[10pt,twocolumn,letterpaper]{article}

\usepackage{iccv}              

\PassOptionsToPackage{table,xcdraw}{xcolor}
\usepackage{xcolor}  

\usepackage{fontawesome}
\usepackage{times}
\usepackage{epsfig}
\usepackage{graphicx}      
\usepackage{amsmath}
\usepackage{amssymb}
\usepackage{booktabs}      
\usepackage{multirow}      
\usepackage{makecell}
\usepackage{etoolbox}
\usepackage[accsupp]{axessibility}  

\usepackage{array}         
\usepackage{colortbl}      

\definecolor{iccvblue}{rgb}{0.21,0.49,0.74}
\usepackage[pagebackref,breaklinks,colorlinks,allcolors=iccvblue]{hyperref}

\def\paperID{660} 
\def\confName{ICCV}
\def\confYear{2025}

\begin{document}
\title{Efficient Multi-Person Motion Prediction by Lightweight Spatial and Temporal Interactions}

\author{
Yuanhong Zheng\textsuperscript{1,3} \quad Ruixuan Yu\textsuperscript{1}\footnotemark[2] \quad Jian Sun\textsuperscript{2,4}\\[0.5em]
{\small $^{1}$Shandong University \quad $^{2}$Xi'an Jiaotong University \quad $^{3}$Peking University \quad $^{4}$Pazhou Laboratory (Huangpu)}\\[0.5em]
{\tt\small zyh0918@mail.sdu.edu.cn \quad yuruixuan@sdu.edu.cn \quad jiansun@mail.xjtu.edu.cn}
}

\maketitle
 
\begin{abstract}
  3D multi-person motion prediction is a highly complex task, primarily due to the dependencies on both individual past movements and the interactions between agents. Moreover, effectively modeling these interactions often incurs substantial computational costs. In this work, we propose a computationally efficient model for multi-person motion prediction by simplifying  spatial and temporal interactions. Our approach begins with the design of lightweight dual branches that learn local and global representations for individual and multiple persons separately. Additionally, we introduce a novel cross-level interaction block to integrate the spatial and temporal representations from both branches. To further enhance interaction modeling, we explicitly incorporate the spatial inter-person distance embedding. With above efficient  temporal and spatial design, we achieve state-of-the-art performance for multiple metrics on standard datasets of CMU-Mocap, MuPoTS-3D, and 3DPW, while significantly reducing the computational cost. Code is available at  \href{https://github.com/Yuanhong-Zheng/EMPMP}{https://github.com/Yuanhong-Zheng/EMPMP}.
\renewcommand{\thefootnote}{\fnsymbol{footnote}}
\footnotetext[2]{Corresponding author.}

\end{abstract}

\section{Introduction}

3D multi-person motion prediction focuses on predicting the future positions of skeletal joints for multiple individuals based on the historical motions.
It has broad applicability across various domains, including human-computer interaction~\cite{azofeifa2022systematic,liu2022arhpe,rautaray2015vision,ren2019human}, virtual/augmented reality~\cite{fu2020capture,clark2020system,ro2019display,rose2018immersion}, sports analysis~\cite{badiola2021systematic,li2021baseball}, and surveillance systems~\cite{calba2015surveillance,hadjkacem2018multi}. 
This task is challenging as it necessitates not only accurate prediction of individual poses through the modeling of historical temporal data, but also a comprehensive representation of inter-person interactions.  

\begin{figure}[t]
\begin{center}
   \includegraphics[width=1.0\linewidth]{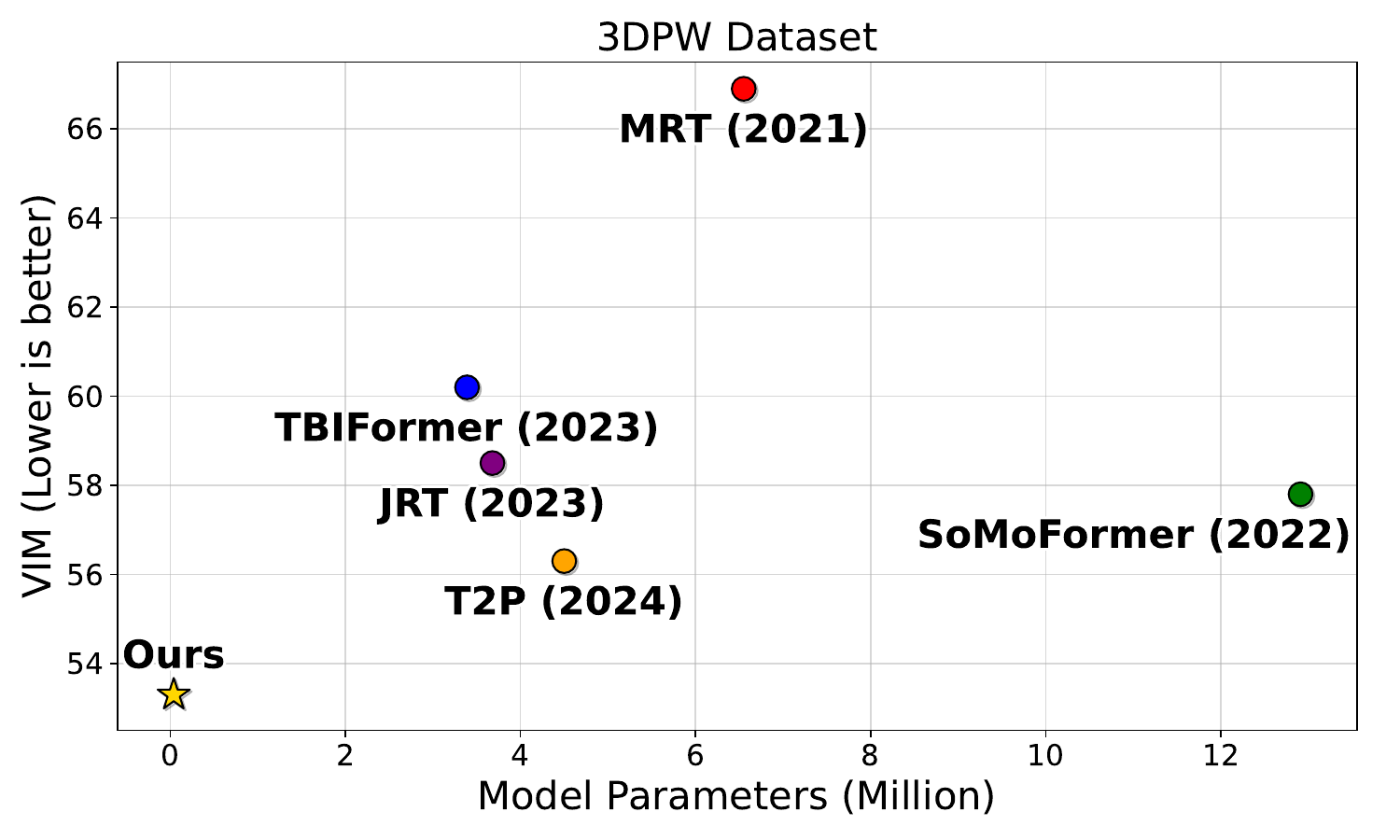}
\end{center}
\setlength{\abovecaptionskip}{-5pt} 
\setlength{\belowcaptionskip}{-10pt} 
\caption{Comparison of our model with  state-of-the-art methods on the 3DPW dataset. Our approach outperforms others while requiring significantly fewer parameters.}
\label{fig:comparison_chart}
\end{figure}

Early studies for 3D motion prediction commonly focused on single-person pose forecasting~\cite{ma2022progressively,tang20233d,mao2020history,mao2019learning,xu2023eqmotion,fragkiadaki2015recurrent,liu2019towards,lehrmann2014efficient}, where the graph neural networks and Transformer-based networks were commonly employed to extract joint-wise features. For example, PGBIG~\cite{ma2022progressively} proposed to alternately extract features through spatial and temporal graph convolutional networks, while STCFormer~\cite{tang20233d} learned cross-joint and cross-frame interaction with Spatial-Temporal Criss-cross attention. Though effective for individual poses, these methods ignored inter-person interactions critical for multi-agent scenarios. Meanwhile, several works on multi-agent trajectory prediction~\cite{zhou2022hivt,amirloo2022latentformer,lee2025mart,chen2023vnagt} have been proposed by treating the agents as points within an interactive system, and conducting trajectory prediction with local context extraction and global interaction modeling. However, these works overlook the intrinsic structural information of individual agents, such as the human skeletal configuration.

Recent research has increasingly focused on multi-person motion prediction, integrating trajectory estimation for global movement and pose forecasting for fine-grained joint-level dynamics~\cite{wang2021multi,peng2023trajectory,xu2023joint,vendrow2022somoformer,jeong2024multi}.  MRT~\cite{wang2021multi} proposed a multi-scale Transformer model with two encoders to capture individual and inter-person interactions  respectively.  TBIFormer~\cite{peng2023trajectory} conducted motion prediction by modeling the body part interactions. JRT~\cite{xu2023joint}  employed  Transformer-based architectures to model both individual joints and inter-person relation, while T2P~\cite{jeong2024multi} separately modeled human pose and trajectory. To the best of our knowledge, all of these methods rely on Transformer-based architectures to capture spatial or temporal interactions. Despite their success in multi-person motion prediction tasks, these models often encounter challenges, including the large number of parameters and the high computational cost associated with Transformer architectures. Furthermore, while they predominantly emphasize the spatial information, there has been limited exploration on the temporal sequential data.


In this work, we propose the Efficient Multi-Person Motion Prediction (EMPMP) network, which leverages a lightweight architecture to effectively capture both spatial and temporal interactions. We first introduce the Multi-level Estimation (ME) block, designed to efficiently extract spatial and temporal features for both individual and inter-person representations. Furthermore, we present the Cross-level Interaction (CI) block, which facilitates the fusion and updating of multi-level features through learned affine transformations. To enhance the modeling of social interactions, we explicitly incorporate inter-person distance into individual representation. Extensive experiments  on standard datasets demonstrate that our approach achieves state-of-the-art performance across multiple evaluation metrics, while utilizing only 1\% to 10\% of the parameters compared to existing methods. As depicted in Fig.~\ref{fig:comparison_chart}, our model outperforms others on the 3DPW dataset, achieving superior performance with a minimal number of parameters.


\section{Related Works}

\noindent\textbf{Single-Person Pose Forecasting.}
The task of single-person pose forecasting aims to predict future human poses based on past observations. Early approaches primarily relied on models such as Hidden Markov Models (HMM)~\cite{lehrmann2014efficient} and Gaussian Processes (GP)~\cite{wang2005gaussian}.  Recurrent Neural Networks (RNNs) were also widely applied for sequential human pose forecasting~\cite{fragkiadaki2015recurrent,liu2019towards,chiu2019action,wang2021pvred}. For example, ERD~\cite{fragkiadaki2015recurrent} combined nonlinear encoders and decoders with RNNs to jointly model human representation and dynamics, while NAFMP~\cite{liu2019towards} utilized Lie algebra and a hierarchical recursive network to capture both local and global contexts. To mitigate error accumulation in frame-by-frame predictions, feed-forward models like Graph Convolutional Networks (GCNs) and Transformer-based architectures were introduced to model long-range dependencies~\cite{mao2019learning,mao2020history,ma2022progressively,dang2021msr}. For instance, LTD~\cite{mao2019learning} proposed learning graph connectivity to capture long-range dependencies in joint sequences, while HRI~\cite{mao2020history} leveraged attention mechanisms to guide predictions based on historical patterns. Despite their success in  pose forecasting tasks, these methods are mainly restricted to single-person scenarios, and they are not suitable for real-world applications involving multiple individuals and inter-person interactions.

\begin{figure*}[!t]
\begin{center}
\includegraphics[width=1.0\linewidth,keepaspectratio]{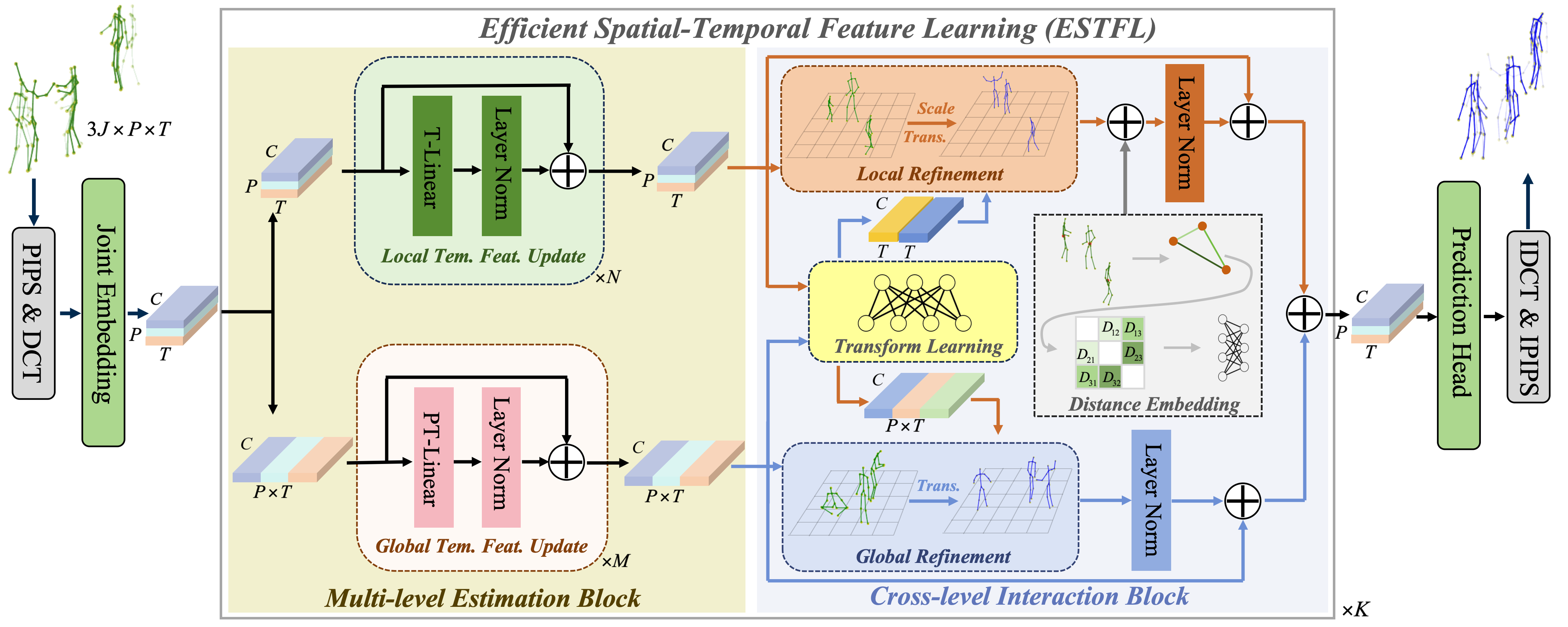}
\end{center}
\setlength{\abovecaptionskip}{-2pt} 
\setlength{\belowcaptionskip}{-5pt} 
\caption{Framework of our EMPMP network. The input is mapped to order-agnostic feature space by PIPS and DCT stages. The Joint Embedding and ESTFL stages are designed to capture spatial and temporal representations as well as their interactions via lightweight architectures. The Prediction Head, IDCT and IPIPS stages are used to map the representation to predicted motion.}
\label{fig:framework}
\end{figure*}


\noindent\textbf{Multi-Person Motion Prediction.} 
This task focuses on predicting the individual movements of people while capturing their complex interactions with others in the scene. Current approaches commonly rely on Transformer-based networks~\cite{wang2021multi,peng2023trajectory,xu2023joint,vendrow2022somoformer,jeong2024multi}, with various design strategies to capture spatial or temporal information. For example, MRT~\cite{wang2021multi} combines local and global encoders to capture both individual and multi-person features, while TBIFormer~\cite{peng2023trajectory} transforms pose sequences into body-semantic-based part sequences for fine-grained interaction modeling. JRT~\cite{xu2023joint} models joint relationships using relative distances and physical constraints. SoMoFormer~\cite{vendrow2022somoformer} represents motion as joint sequences and encodes joint types, identities, and global positions. T2P~\cite{jeong2024multi} decouples the task into local pose and global trajectory predictions.
While these methods effectively model interactions, they often require high parameter counts and computational resources. In this work, we propose a novel lightweight network to efficiently captures spatial and temporal representations as well as their interactions, achieving state-of-the-art results with significantly lower computational cost.


\noindent\textbf{MLP-based Lightweight Neural Networks.}
It is crucial to maintain a lightweight architecture while ensuring high performance. Conventional approaches, such as model pruning~\cite{shim2021layer}, knowledge distillation~\cite{liu2022transkd}, and sparse attention~\cite{child2019generating}, have been widely adopted. MetaFormer~\cite{yu2022metaformer} showed that the Transformer’s strength primarily lies in its overall design rather than Self-Attention, motivating MLP-based alternatives. MLP-Mixer~\cite{tolstikhin2021mlp} and CycleMLP~\cite{chen2021cyclemlp} replace attention with linear layers, achieving competitive performance with fewer parameters. For the single-person 3D pose forecasting task, Motion Mixer~\cite{bouazizi2022motionmixer} captures spatial and temporal dependencies with separate MLPs, while SiMLPe~\cite{guo2023back} integrates DCT~\cite{ahmed1974discrete} and cascaded MLPs to strengthen temporal representations. GraphMLP~\cite{li2025graphmlp} combines MLPs and GCN in a unified framework for 3D pose forecasting from 2D images or historical motions. In this work, we integrate MLPs into 3D multi-person motion prediction for the first time, and propose novel lightweight yet effective local-global and spatial-temporal interaction operations for enhanced representation learning.

\section{Method}
\label{sec3:Architecture}

Given the historical 3D observation of multi-person joint sequences ${X}=\{  {x}_{t}^{p}\}_{1:T}^{1:P} \in \mathbb{R}^{3J\times P\times T}$, which consists of $P$ persons, $T$ time frames, and $J$ joints per person, our goal is to predict the future joint sequences ${  {Y}}=\{  {y}_{t}^{p}\}_{1:T'}^{1:P}$ for the next $T'$ frames. 
In this paper, we propose the Efficient Multi-Person Motion Prediction (EMPMP) network with lightweight yet expressive spatial and temporal interactions for 3D multi-person motion prediction. The framework is presented in Sects.~\ref{subsec3.1:Overall Framework}-~\ref{subsec3.4:Trajectory Prediction}, with network training and implementation details in Sect.~\ref{subsec3.5:Network Training and Implementation Details}.


\subsection{Overall Framework}
\label{subsec3.1:Overall Framework}
Our EMPMP network conducts 3D multi-person motion prediction through a cascaded pipeline as shown in Fig.~\ref{fig:framework}. The framework comprises three core modules: the Joint Embedding for encoding skeletal geometry, the Efficient Spatial-Temporal Feature Learning to model multi-level spatial and temporal interactions, and the Prediction Head for future motion prediction mapping, as detailed in the subsequent subsections.

To resolve order ambiguity of individuals, we introduce Permutation-Invariant Person Sorting (PIPS), which reorders individuals by descending pairwise distances in the first frame, yielding order-agnostic features  $X'\in \mathbb{R}^{3J\times P\times T}$ (please refer to supplementary material for details). Following this, we apply Discrete Cosine Transform (DCT) to compress temporal dynamics into a low-frequency spectrum ${X''} = D X'$, where $D \in \mathbb{R}^{T \times T}$ is the  DCT  coefficient matrix conducted on temporal dimension. 





\subsection{Joint Embedding}
\label{subsec3.2:Joints Embedding}
The joint embedding stage transforms the DCT-processed input ${X''}$   into a high-dimensional feature space through a linear projection, integrating spatial joint information into a structured representation: 
\begin{equation} 
Z = {X''} W_0 + b_0, \quad Z \in \mathbb{R}^{C \times P \times T}, 
\end{equation} 
where $W_0 \in \mathbb{R}^{3J \times C}$ and $b_0 \in \mathbb{R}^{C}$ are learnable parameters shared across all individuals ($P$) and frames ($T$).
This embedding integrates spatial features into a shared feature space, enabling holistic pose modeling rather than isolated joint processing.


\subsection{Efficient Spatial-Temporal Feature Learning}
\label{subsec3.3:MP-MLP Stage}

The Efficient Spatial-Temporal Feature Learning (ESTFL) stage forms the core of our framework, designed to capture inter-person spatial dependencies and inter-frame temporal dynamics within the sequential multi-person representation $Z\in \mathbb{R}^{C\times P\times T}$. It comprises two main  components: the \textit{Multi-level Estimation} (ME) block  and \textit{Cross-level Interaction} (CI) block, described as follows.

\subsubsection{Multi-level Estimation (ME) Block}
\label{subsec3.3:Local&Global Block}

The ME block is proposed to capture temporal information for both individual and multi-person representations. A dual-stream  architecture is employed to capture both local (individual) and global (scene-level) dynamics.

\noindent\textbf{Local Temporal Feature Update.}
The local stream focuses on capturing individual motion patterns  through iterative temporal feature updates. In each iteration, a shared temporal linear transformation (T-linear) is applied across all persons, followed by layer normalization and residual connection. Formally, the $i$-th iteration of local update  is:
\begin{equation} 
Z_l^i = Z_l^{i-1} + \beta (Z_l^{i-1} W_l^i + b_l^i), \quad i = 1, ..., N, 
\end{equation} 
where $Z_l^0 = Z$,  $W_l^i \in \mathbb{R}^{T \times T}$ and $b_l^i \in \mathbb{R}^{T}$ are learnable parameters, and $\beta$ denotes layer normalization.
By sharing these parameters across all persons, this iterative process progressively refines fine-grained individual motion understanding while preserving computational efficiency.

\noindent\textbf{Global Temporal Feature Update.} 
The global stream is designed to learn temporal interactions across all individuals within the scene. We merge the person  ($P$) and temporal ($T$) dimensions of $Z$ to form a global representation $Z_g^0\in \mathbb{R}^{C\times PT}$, which encodes dynamics of the entire scene. A linear transformation along the person-temporal dimension (PT-linear) is applied  with a residual connection, and the global features are iteratively refined  by:
\begin{equation} 
Z_g^j = Z_g^{j-1} + \gamma (Z_g^{j-1} W_g^j + b_g^j), \quad j = 1, ..., M, 
\end{equation} 
where $W_g^j \in \mathbb{R}^{PT \times PT}$ and $b_g^j \in \mathbb{R}^{PT}$ are learnable parameters for the $j$-th iteration, and $\gamma$ is layer normalization.  By progressively refining the global representation, the global stream effectively captures the temporal dependencies across all individuals in the scene.

The multi-level features of  ${Z}_{l}^N \in \mathbb{R}^{C\times P \times T}$ and ${Z}_{g}^M \in \mathbb{R}^{C \times PT}$ are subsequently integrated in the Cross-level Interaction (CI) block, enabling a comprehensive interaction between individual motion details and global scene dynamics for enhanced motion prediction.

\subsubsection{Cross-level Interaction (CI) Block  }
\label{subsec3.4:LGI Block}
The CI block facilitates efficient information exchange between local and global representations. For simplicity, we omit the superscripts and denote the local and global features as $Z_l$ and $Z_g$, respectively.

\noindent\textbf{Local Refinement with  Global Representation.} 
The global representation $Z_g$ captures high-level temporal interactions across all individuals in the scene, providing essential contextual information to enhance  local features of each individual.
To refine local representations, we apply affine transformations with  scale  $  {S}\in \mathbb{R}^{C\times T}$ and translation  $  {H}\in \mathbb{R}^{C\times T}$ learned from $Z_g$ via linear transformations:
\begin{equation}
  {S} =   {Z}_{g}  {W}_S+  {b}_S, \quad 
  {H} =   {Z}_{g}  {W}_H+  {b}_H, 
\end{equation} 
where $  {W}_S,  {W}_H\in \mathbb{R}^{PT \times T}$ and $  {b}_S,  {b}_H\in \mathbb{R}^{ T}$ are learnable parameters, and they are employed to   integrate global context across individuals and temporal frames. To maintain consistent global influence across all individuals, $  {S}$ and $  {H}$ are shared across persons,  yielding expanded representations  $\Hat{  {S}}, \Hat{  {H}} \in \mathbb{R}^{C \times P\times T}$. The refined local representation is then computed by applying these transformations:
\begin{equation}
  {Z}_{l}' = \rho(  {Z}_{l} \odot (1 + \Hat{  {S}}) + \Hat{  {H}}),
\end{equation}
where $ \odot$  denotes the Hadamard (element-wise) product, and $\rho$ is layer normalization. This process dynamically adjusts  local features using  global context, ensuring that the refined representations capture both individual motion details and scene-level temporal dynamics.

\noindent\textbf{Inter-person Distance Embedding.}
To further refine the local representation, we incorporate the spatial relationships between individuals in the scene by computing an inter-person distance matrix $  {D} \in \mathbb{R}^{P \times P \times T}$,  which captures the spatial interactions between individuals at each time frame. The matrix element is defined as
\begin{equation}
    D_{p_1p_2t} = \|{x}_{t,\textbf{hip}}^{p_1} - {x}_{t,\textbf{hip}}^{p_2}\|_2,
\end{equation}
where $D_{p_1p_2t}$ represents the Euclidean distance between the hips of the $p_1$-th and $p_2$-th persons  at time $t$.
This distance matrix is then linearly transformed along its first dimension to produce the distance embedding:
\begin{equation}
\tau(  {D}) =   {D}  {W}_D+  {b}_D, 
\end{equation} 
where $W_D\in \mathbb{R}^{P \times C}$ and $b_D\in \mathbb{R}^{C}$ are learnable parameters. The embedded distance $\tau(D) \in \mathbb{R}^{C \times P \times T}$ is then used to refine the local representation:
\begin{equation}
      {Z}_{l}^* =   {Z}_{l} + \xi(  {Z}_{l}' + \tau(  {D})),
\end{equation} 
where $\xi$ represents layer normalization.
This process ensures that the local representation incorporates both individual motion dynamics and the spatial relationships between individuals, thereby improving the model's accuracy in predicting multi-person motions.

\noindent\textbf{Global Refinement with Local Representation.}
To refine the global representation using local information from $  {Z}_{l} \in \mathbb{R}^{C \times P\times T}$, we first compress the local features across individuals and time into a compact representation for affine transformation. Specifically, we combine the  $P$ and $T$ dimensions of $  {Z}_{l}$, resulting in $  {Z}_{l2g}\in \mathbb{R}^{C \times PT}$, and  apply a linear transformation to learn the global translation: 
\begin{equation}
    {G} =   {Z}_{l2g}  {W}_G+  {b}_G, 
\end{equation}
where $  {W}_G\in \mathbb{R}^{PT\times PT},  {b}_G\in \mathbb{R}^{PT}$ are learnable parameters.  Based on above learned translation, the refined global representation is computed as
\begin{equation}
    {Z}_g^* =   {Z}_{g} + \pi(  {Z}_g +   {G}),   
\end{equation}
where $\pi$  is layer normalization. This process  ensures that the global representation is enriched with detailed local dynamics while maintaining stable feature distributions.

Finally, the refined local and global representations are combined into a unified feature:
\begin{equation}
  {Z^*} ={  {Z}}_{l}^*+\alpha\hat{  {Z}}_{g}^*,~~~  {Z^*}\in \mathbb{R}^{C \times P \times T},
  \label{eq:combine}
\end{equation}
where $\hat{.}$ denotes dimension transform for compatibility, and 
$\alpha$ is the combination parameter.

The CI block enables bidirectional information exchange between local and global representations, significantly improving the model's ability to capture complex motion dynamics while ensuring computational efficiency.

The ESTFL stage employs a dual-block architecture, consisting of the ME block for multi-level feature learning and the CI block for cross-level information exchange. This straightforward yet effective design allows the model to capture multi-level spatial and temporal interactions, boosting its ability to predict motion dynamics. Experimental results demonstrate the superior performance of above designs, showing substantial improvements in both accuracy and efficiency for 3D multi-person motion prediction task.

\subsection{Prediction Head and Loss Function}
\label{subsec3.4:Trajectory Prediction}
Given  $Z^*\in \mathbb{R}^{C\times P \times T}$, the prediction head is designed to map the high-dimensional features to joint-wise representations through linear transformation in two dimensions:
\begin{equation}
  \tilde{Z}=   {Z^*}  {W}_T+  {b}_T,~~~\hat{Z}=   {\tilde{Z}}  {W}_C+  {b}_C,
\end{equation}
where $W_T\in \mathbb{R}^{T\times T'},b_T\in \mathbb{R}^{T'} $ and $W_C\in \mathbb{R}^{C\times 3J},b_C\in \mathbb{R}^{3J} $ are learnable parameters that perform the transformations in temporal ($T$) and feature ($C$) dimensions, respectively. The final joint sequence is obtained by applying  Inverse  DCT  cross the temporal
dimension as ${{Y'}}={D}^{-1}\hat{Z}$, followed by Inverse PIPS to restore the order and generate a permutation-invariant prediction $Y\in \mathbb{R}^{3J\times P \times T'}$.

The loss function consists of two components including  the mean joint loss and the velocity loss as utilized in~\cite{guo2023back}:
\begin{equation} 
\mathcal{L} = \mathcal{L}^{Joint} + \mathcal{L}^{Vel}. 
\end{equation}
The mean joint loss quantifies the discrepancy between the predicted and ground-truth joint sequences by:
\begin{equation} 
\mathcal{L}^{Joint} = \frac{1}{P \cdot T' \cdot J} \sum_{p=1}^{P} \sum_{t=1}^{T'} \sum_{j=1}^{J} \|   {y}_t^{p,j} - \hat{{y}}_t^{p,j} \|^2,
\end{equation} 
where ${  {y}}_t^{p,j} \in \mathbb{R}^{3}$ represents the predicted world coordinates of the $j$-th joint for the $p$-th person at the $t$-th frame, and $\hat{  {y}}_t^{p,j}$ is the corresponding ground truth.
The velocity loss accounts for the predicted and ground-truth velocities between consecutive frames:
\begin{equation} 
\mathcal{L}^{Vel} = \frac{1}{P \cdot (T'-1) \cdot J} \sum_{p=1}^{P} \sum_{t=1}^{T'-1} \sum_{j=1}^{J} \| v_t^{p,j} - \hat{v}_t^{p,j} \|^2, \end{equation} 
where the predicted and ground-truth velocities are computed as follows:
\begin{equation}
 v_t^{p,j} = {  {y}}_{t+1}^{p,j} - {  {y}}_{t}^{p,j}, \quad \hat{v}_t^{p,j}  =   \hat{y}_{t+1}^{p,j} -   \hat{y}_{t}^{p,j}.
\end{equation}
By incorporating both joint and velocity constraints, the  loss function enforces consistency in both spatial and temporal dynamics, thereby enhancing the accuracy and realism of the motion prediction.

\subsection{Network Training and Implementation Details}
\label{subsec3.5:Network Training and Implementation Details}
 Our network is built on the framework proposed in  Sects.~\ref{subsec3.1:Overall Framework}-~\ref{subsec3.4:Trajectory Prediction}. For the hyper-parameters, we set the combination parameter $\alpha=0.2$ in Eq.~(\ref{eq:combine}), and we take $K=4$ stages of ESTFL to efficiently capture the spatial and temporal features. Each ESTFL stage consists of $N=16, M=1$ iterations for local and global temporal feature updates, respectively. The feature channel dimension is set to  $C=45$ for  experiments on CMU-Mocap and MuPoTs-3D datasets, and $C=39$ for experiments on the 3DPW dataset.

The network is trained with batch size  128 and  learning rate  $3 \times 10^{-4}$ by Adam optimizer.  During training, we apply  data augmentation after PIPS stage, with random vertical-axis rotations and person order permutations  as~\cite{vendrow2022somoformer}. All experiments are conducted on a single NVIDIA 3090 GPU. Our code will be released if the paper is accepted.


\section{Experiments}
\label{sec4:Experiments}
In this section, we first present the dataset settings and evaluation metrics, followed by comparisons of the experimental results. Additionally, we conduct ablation studies to  investigate the effectiveness of our designs.

\begin{table*}[ht]
    \centering
    \small
    \renewcommand{\arraystretch}{1.0} 
    \resizebox{\textwidth}{!}{ 
    \begin{tabular}{>{\centering\arraybackslash}p{0.5cm}l|>{\centering\arraybackslash}p{1.4cm}|>{\centering\arraybackslash}p{1.4cm}|>{\centering\arraybackslash}p{1.2cm}|>{\centering\arraybackslash}p{1.2cm}|>{\centering\arraybackslash}p{1.4cm}|>{\centering\arraybackslash}p{1.4cm}|>{\centering\arraybackslash}p{1.2cm}|>{\centering\arraybackslash}p{1.2cm}}
        \toprule
        \midrule
        \multirow{3.5}{*}{\rotatebox{90}{Metric}}
        &\multicolumn{1}{l|}{\multirow{2}{*}{Settings}}         & \multirow{2}{*}{3DPW-Ori} &\multirow{2}{*}{3DPW-RC}  & \multicolumn{2}{c|}{\multirow{2}{*}{CMU-Syn}} &{AMASS\slash3DPW-Ori} &AMASS\slash3DPW-RC & \multicolumn{2}{c}{\multirow{2}{*}{CMU-Syn/MuPoTS}}   \\ \cmidrule{3-10}
        &\multicolumn{1}{l|}{In/Out Length}   &\multicolumn{2}{c|}{1030ms / 900ms}   & \multicolumn{1}{c|}{2s / 2s} & 1s / 1s &\multicolumn{2}{c|}{1030ms / 900ms} & \multicolumn{1}{c|}{2s / 2s} & 1s / 1s              \\ \midrule
        \multirow{6}{*}{\rotatebox{90}{\textbf{MPJPE}}} 
                & MRT~\cite{wang2021multi}{\scriptsize{$^{\rm \textcolor{blue}{'2021}}$}}              &167.1	&134.2	&160.7	&83.5	&145.3	&124.8	&181.0	&93.1                             \\
                               & SoMoFormer~\cite{vendrow2022somoformer}{\scriptsize{$^{\rm \textcolor{blue}{'2022}}$}}         & 150.0        & 107.7            & 148.3  & 77.0 &\textbf{114.6}&96.4&162.0&\textbf{89.1}                        \\
                               & TBIFormer~\cite{peng2023trajectory}{\scriptsize{$^{\rm \textcolor{blue}{'2023}}$}}          & 149.9          & 124.1            & 169.7   & 91.0   &140.4&114.3&173.4&93.2                        \\
                               & JRT~\cite{xu2023joint} {\scriptsize{$^{\rm \textcolor{blue}{'2023}}$}}               & 154.9          & 114.0               & 157.6    & 84.3  &131.9&102.3&161.5&93.0                          \\ 
                               & T2P~\cite{jeong2024multi}{\scriptsize{$^{\rm \textcolor{blue}{'2024}}$}}      & 137.5          & 110.1        & 136.4    & 75.9  &121.1&100.4&163.6& 94.7                         \\ 
        \rowcolor{gray!20}     & Ours              & \textbf{131.8}          & \textbf{99.5}              & \textbf{128.0}                      & \textbf{73.5} &119.9&\textbf{92.1}&\textbf{160.1}&92.4                         \\ \midrule
        
        \multirow{6}{*}{\rotatebox{90}{\textbf{VIM}}} 
        & MRT~\cite{wang2021multi}{\scriptsize{$^{\rm \textcolor{blue}{'2021}}$}}   &66.9	&55.2	&61.5	&36.3	&59.2	&52.3	&70.1	&41.6                                      \\
                               & SoMoFormer~\cite{vendrow2022somoformer}{\scriptsize{$^{\rm \textcolor{blue}{'2022}}$}}         & 57.8          & 44.4      & 56.7   & 34.4    &\textbf{46.3}&40.0&63.7&\textbf{39.3}                                  \\
                               & TBIFormer~\cite{peng2023trajectory}{\scriptsize{$^{\rm \textcolor{blue}{'2023}}$}}          & 60.2           & 51.2       & 64.6     & 40.3  &56.4&47.4&68.3&44.0                 \\
                               & JRT~\cite{xu2023joint}{\scriptsize{$^{\rm \textcolor{blue}{'2023}}$}}                 & 58.5           & 47.0      & 56.6     & 34.0  &47.2&39.5&64.7&41.2                           \\ 
                                & T2P~\cite{jeong2024multi}{\scriptsize{$^{\rm \textcolor{blue}{'2024}}$}}            & 56.3           & 46.9      & 54.6   & 34.2  &49.4&42.4&64.0&42.0                            \\ 
        \rowcolor{gray!20}     & Ours              & \textbf{53.3}           & \textbf{41.3}      & \textbf{50.2}                      & \textbf{32.4}   &48.6&\textbf{38.4}&\textbf{62.9}&40.6                              \\ \midrule
        
        \multirow{6}{*}{\rotatebox{90}{\textbf{APE}}} 
        & MRT~\cite{wang2021multi}{\scriptsize{$^{\rm \textcolor{blue}{'2021}}$}}         &125.1	&123.6	&95.5	&60.0	&115.3	&110.8	&135.1	&81.2 \\
                               & SoMoFormer~\cite{vendrow2022somoformer}{\scriptsize{$^{\rm \textcolor{blue}{'2022}}$}}          & 118.3          & 114.9      & 87.1   &55.2&97.5&101.4&130.8&80.6                                       \\
                               & TBIFormer~\cite{peng2023trajectory}{\scriptsize{$^{\rm \textcolor{blue}{'2023}}$}}           & 115.9           & 115.5     &96.9 &63.6 &110.3 &108.0  & 137.7     & 84.4                 \\
                               & JRT~\cite{xu2023joint} {\scriptsize{$^{\rm \textcolor{blue}{'2023}}$}}               & 123.0           & 120.5      & 95.8     &61.6   &114.1 &112.5 &125.7 & 78.4                           \\ 
                                & T2P~\cite{jeong2024multi}{\scriptsize{$^{\rm \textcolor{blue}{'2024}}$}}            & 115.1           & 115.1      & 94.5   &59.9     &110.1 &108.7 &147.7 &92.8                          \\ 
        \rowcolor{gray!20}     & Ours              & \textbf{98.6}           & \textbf{96.6}      & \textbf{83.2}                      & \textbf{52.3}  &\textbf{95.0}  &\textbf{90.6} &\textbf{124.9} &\textbf{76.2}                              \\ \midrule
        \bottomrule
    \end{tabular}
    }
    \setlength{\abovecaptionskip}{8pt} 
    \setlength{\belowcaptionskip}{-10pt} 
    \caption{Performance comparisons across multiple settings using MPJPE, VIM, and APE metrics. Our method achieves superior performance with significantly fewer parameters. The best result is highlighted in bold.}
    \label{tab:main-table}
\end{table*}

\subsection{Datasets and Evaluation Metrics}
\label{subsec4.2:Datasets}
\noindent\textbf{3DPW and AMASS.} The 3DPW dataset~\cite{von2018recovering} is a large-scale collection containing over 51,000 frames of human activities in unconstrained environments, with scenes involving two individuals. The AMASS dataset~\cite{mahmood2019amass} is a comprehensive motion capture dataset, featuring over 40 hours of motion data from 300+ subjects performing more than 11,000 distinct motions. These datasets are used in four distinct settings:
1)~\textit{3DPW-Ori}: As in~\cite{vendrow2022somoformer, xu2023joint}, the input consists of 1030ms (16 frames) and the prediction horizon is 900ms (14 frames) based on the original 3DPW dataset.
2)~\textit{3DPW-RC}: This setting is similar to~\textit{3DPW-Ori}, but with camera displacements removed to enhance the realism of human motion, as in~\cite{xu2023joint}.
3)~\textit{AMASS/3DPW-Ori}: The model is pre-trained on AMASS, then fine-tuned and evaluated on~\textit{3DPW-Ori}, with pre-training and fine-tuning configurations matching those in ~\cite{vendrow2022somoformer, xu2023joint}.
4)~\textit{AMASS/3DPW-RC}: Similar to \textit{AMASS/3DPW-Ori}, but the model is fine-tuned and evaluated on \textit{3DPW-RC} instead, as in~\cite{xu2023joint}.

\noindent\textbf{CMU-Mocap.}  The CMU-Mocap dataset~\cite{cmu-mocap} consists of 140 subjects and 2,605 sequences featuring one or two individuals. Following~\cite{wang2021multi}, 6,000 synthetic scenes with three individuals are generated, each consisting of 60 frames at a frame rate of 15 FPS. The \textit{CMU-Syn} setting includes three motion prediction tasks: predicting 1 second of motion from  1-second input (1s/1s) for short-term prediction, predicting 2 seconds from 2-second input (2s/2s) for long-term prediction, and predicting 3 seconds from  1-second input (1s/3s) for extended forecasting. For the 1s/1s setting, we randomly sample 30 frames from 60 during training, increasing the training samples to enhance model robustness.


\noindent\textbf{MuPoTS-3D.} The MuPoTS-3D dataset~\cite{mehta2018single} contains over 8,000 frames across 20 real-world scenes with up to three subjects. Similar to strategy in~\cite{wang2021multi}, for the \textit{CMU-Syn/ MuPoTS} setting, the models are trained on \textit{CMU-Syn}, and performance is evaluated by predicting 15 frames (1s) and 30 frames (2s) of motion sequences on  MuPoTS-3D dataset.

\noindent\textbf{Evaluation Metrics.}
\label{subsec4.1:Metrics}
We  evaluate  model performance using metrics of Mean Per Joint Position Error (MPJPE)~\cite{ionescu2013human3}, Visibility-Ignored Metric (VIM)~\cite{adeli2021tripod} and Aligned Mean Per Joint Position Error (APE).  MPJPE measures the mean Euclidean distance between predicted and ground truth 3D joint positions across all frames. VIM assesses the per-frame accuracy by computing  mean joint-coordinate distance. APE removes global displacement, focusing on local posture prediction. For these metrics, lower values indicate better performance. Due to space constraints, we report average values for key frames in APE and VIM. Further details on selection scheme and additional metrics are available in the supplementary material.

\subsection{Experimental Results}
\label{sec5:Experimental Results}
We evaluate our approach on 3D multi-person motion prediction tasks in three scenarios: standard, incorporating pre-training, and cross-dataset, to comprehensively assess the model's effectiveness and generalizability. 


\noindent\textbf{Results on \textit{3DPW-Ori} and \textit{3DPW-RC}.}
\label{subsec5.1:Results on 3DPW and 3DPW/RC}
As shown in the first and second columns of Tab.~\ref{tab:main-table}, our model outperforms previous methods in both \textit{3DPW-Ori} and \textit{3DPW-RC} settings for standard multi-person motion prediction tasks, demonstrating significant superiority. The substantial reductions in MPJPE, VIM, and APE highlight the effectiveness of our approach. Notably, our model achieves state-of-the-art performance with only 0.04M parameters, as shown in Tab.~\ref{tab:comparison_updated}, which constitutes less than approximately 1.2\% of the parameter count of prior models. Additionally, our model requires less GPU memory and computational FLOPs, and more detailed comparisons on computational costs can be found in the supplementary material.

\begin{table}[ht]
    \centering
    \renewcommand{\arraystretch}{1.0} 
    \resizebox{\linewidth}{!}{ 
    \begin{tabular}{l|>{\centering\arraybackslash}p{1.5cm}|>{\centering\arraybackslash}p{1.5cm}|>{\centering\arraybackslash}p{1.5cm}|>{\centering\arraybackslash}p{1.5cm}}
        \toprule
        \midrule     
        \multicolumn{1}{l|}{\multirow{2}{*}{Settings}}    & \multicolumn{1}{c|}{\multirow{2}{*}{3DPW}} & 3DPW-Pretrain      & \multicolumn{2}{c}{\multirow{2}{*}{CMU-Syn \& MuPoTS}}     \\ \cmidrule{2-5}         
        \multicolumn{1}{l|}{In/Out Length}  & \multicolumn{2}{c|}{1030ms / 900ms} & 2s / 2s & 1s / 1s  \\ \midrule

         MRT~\cite{wang2021multi}                   & 6.55M             & 6.55M   &  6.29M &  6.29M                  \\
                                SoMoFormer~\cite{vendrow2022somoformer}                    & 12.91M         & 12.91M      &12.91M  & 12.91M                                \\
                                TBIFormer~\cite{peng2023trajectory}               & 3.39M                        & 3.39M       & 7.31M &  4.24M                 \\
                                JRT~\cite{xu2023joint}                        & 3.68M                     & 3.68M & 3.68M & 3.68M                   \\ 
                                T2P~\cite{jeong2024multi}        & 4.50M    & 4.50M & 4.60M &  4.50M  \\ 
                               
        \rowcolor{gray!20}      Ours                     & \textbf{0.04M}                     & \textbf{0.65M}   & \textbf{0.17M} & \textbf{0.05M}                  \\ \midrule
        \bottomrule
    \end{tabular}
    }
    \setlength{\abovecaptionskip}{8pt} 
    \setlength{\belowcaptionskip}{-5pt} 
    \caption{Model Parameters (in Millions) for different settings. 3DPW refers to the settings of \textit{3DPW-Ori} and \textit{3DPW-RC}, 3DPW-Pretrain covers \textit{AMASS/3DPW-Ori} and \textit{AMASS/3DPW-RC}, while CMU-Syn\&MuPoTS include \textit{CMU-Syn} and \textit{CMU-Syn/MuPoTS}.}
    \label{tab:comparison_updated}
\end{table}

\noindent\textbf{Results on \textit{CMU-Syn}.}
\label{subsec5.2:Results on CMU-Mocap}  As demonstrated in the third and fourth columns of Tab.~\ref{tab:main-table}, our model outperforms the compared methods across all three evaluation metrics for both short-term (1s/1s) and long-term (2s/2s) motion prediction in the \textit{CMU-Syn} setting, while maintaining a lightweight design with less parameters as shown in Tab.~\ref{tab:comparison_updated}. Furthermore, the results for the extended forecasting task (1s/3s) are presented in Tab.~\ref{tab:Mocap1sto3s}, where our EMPMP consistently surpasses the other methods within the 3s prediction horizon, particularly for long-range predictions at 2s and 3s.

\noindent\textbf{Results on \textit{AMASS/3DPW-Ori} and \textit{AMASS/3DPW-RC} with Pre-training.}
\label{subsec5.4:Results with Pre-training}
For tasks involving a pre-training strategy, we utilize a 0.65M-parameter model to enhance our modeling capability, ensuring a fair comparison (please refer to the supplementary material for further details). As shown in Tab.~\ref{tab:main-table}, our model achieves state-of-the-art performance in four out of six comparisons, while remaining highly competitive in the other two.

\begin{figure*}[!t]
\begin{center}
\includegraphics[width=1\linewidth,height=9cm]{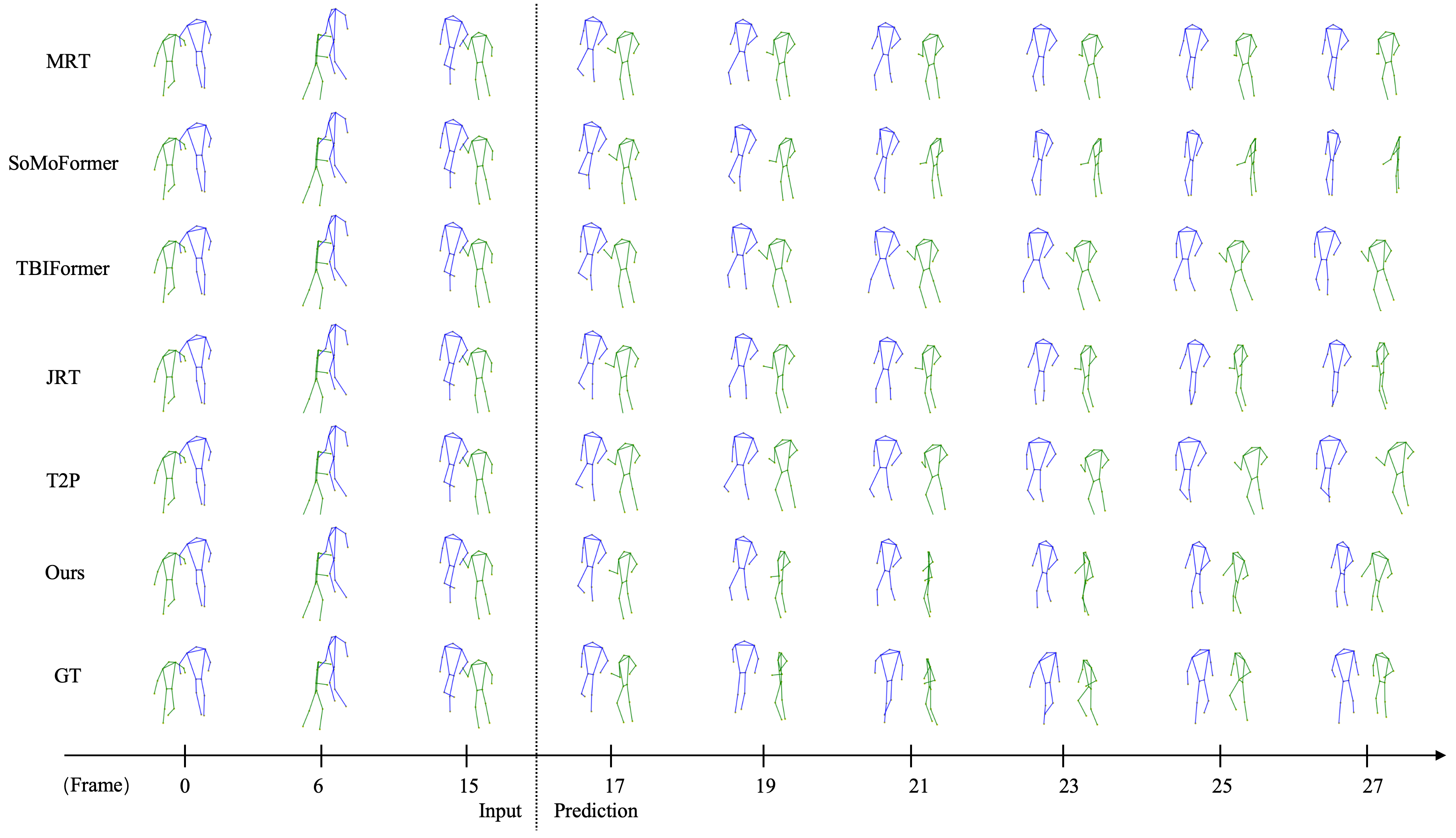}
\end{center}
\setlength{\abovecaptionskip}{-2pt} 
\setlength{\belowcaptionskip}{-5pt} 
\caption{Qualitative results in the \textit{3DPW-RC} setting involve predicting 14 frames using 16 frames of input. Different colors are used to distinguish between individuals in the sequence. }
\label{fig:visualization}
\end{figure*}

\noindent\textbf{Results on \textit{CMU-Syn/MuPoTS} for Cross-Dataset Evaluation.}
As shown in the last two columns of Tab.~\ref{tab:main-table}, our model excels in cross-dataset evaluation, achieving the best performance in the 2s/2s setting while using fewer parameters. For the 1s/1s setting, our model attains the best APE and competitive MPJPE/VIM results, demonstrating strong generalization capabilities.

\begin{table}[]
	\centering
        \Large
        \resizebox{\columnwidth}{!}{
	\begin{tabular}{l|ccc|ccc|ccc}
            \toprule
            \midrule     
		  Metric      & \multicolumn{3}{c|}{MPJPE}        & \multicolumn{3}{c|}{VIM}       & \multicolumn{3}{c}{APE}        \\ \cmidrule{2-10}
		Out Length  & 1s       & 2s       & 3s        & 1s       & 2s       & 3s       & 1s       & 2s       & 3s       \\ \midrule
		MRT~\cite{wang2021multi}       & 89	&145	&189	&65 	&97	&128	&86	&102	&111        \\
		SoMoFormer~\cite{vendrow2022somoformer}  & \textbf{84}          &  151         &   202        &  66      &  108     &    140    &   81    &   101    &  112     \\
		TBIFormer~\cite{peng2023trajectory}   & 98         &    165       &    218       &   72       &  114    &   148     &  88        &   103       &  116        \\
		JRT~\cite{xu2023joint}         & 88          &   154        &   206        & \textbf{60}        &  105        & 147 &  89  & 106 &  118 \\
		T2P~\cite{jeong2024multi}         &  89         &  139         &   177        & 65 &   89  &   117  &  87 &   103  &   116  \\
\rowcolor{gray!20} Ours       & \textbf{84}          & \textbf{130}          & \textbf{167} &   \textbf{60}   &   \textbf{85}   &   \textbf{113}  & \textbf{77}      &  \textbf{93}     & \textbf{103}    \\ \midrule
        \bottomrule
	\end{tabular}
    }
    \setlength{\abovecaptionskip}{8pt} 
    \setlength{\belowcaptionskip}{-10pt} 
	\caption{Performance comparison  in \textit{Mocap-Syn} (1s/3s) setting. }
	\label{tab:Mocap1sto3s}
\end{table}

\subsection{Ablation Study}
\label{subsec5.5:Ablation Study}
In this subsection, we conduct detailed ablation studies to assess  the effectiveness of our designs and the impact of hyper-parameters.  All of the models are evaluated  by MPJPE in the \textit{3DPW-Ori} and \textit{CMU-Syn} settings.

\noindent\textbf{Effectiveness of Multi-level Estimation.}
The ME block integrates individual and multi-person features via local and global temporal feature update mechanisms. As shown in the first and second rows of Tab.~\ref{tab:Key Components}, removing either the local temporal feature update (LTFU) or the global temporal feature update (GTFU) stream results in a performance decline compared to the network that includes both components, as seen in the third row. These comparisons underscore the importance of both streams and highlight the effectiveness of the proposed ME block.

\noindent\textbf{Effectiveness of Cross-level Interaction.}
The CI block enhances local and global representations through interaction via learned affine transformations (LRwGR and GRwLR). As shown in Tab.~\ref{tab:Key Components}, removing the entire CI block (third row), or individually removing either the local or global refinements (fourth and fifth rows), significantly reduces performance compared to the model with CI block (last two rows). This clearly validates the critical role of each component. Notably, we learn both scale and translation parameters during local refinement, whereas global refinement employs only translation. Further ablation studies in the supplementary material confirm the effectiveness of these design choices in managing affine transformations.

\begin{table}[ht]
    \centering
    \large
    \renewcommand{\arraystretch}{1} 
    \setlength{\tabcolsep}{8pt} 
    \resizebox{\linewidth}{!}{ 
    \begin{tabular}{ccccc|c}
        \toprule
        \textbf{LTFU} & \textbf{GTFU} & \textbf{LRwGR} & \textbf{GRwLR} & \textbf{DE} & \textbf{MPJPE} \\ 
        \midrule
        × & \checkmark & × & × & × & 160.4 (0.04M) \\ 
        \checkmark & × & × & × & × & 143.4 (0.07M) \\ 
        \checkmark & \checkmark & × & × & × & 140.9 (0.10M) \\ 
        \checkmark & \checkmark & × & \checkmark & × & 137.7 (0.12M) \\ 
        \checkmark & \checkmark & \checkmark & × & × & 133.8 (0.14M) \\ 
        \checkmark & \checkmark & \checkmark & \checkmark  & × & 129.8 (0.16M) \\ 
        \checkmark & \checkmark & \checkmark & \checkmark & \checkmark  & \textbf{128.0 (0.17M)} \\ 
        \bottomrule
    \end{tabular}
    }
        \setlength{\abovecaptionskip}{8pt} 
        \setlength{\belowcaptionskip}{-10pt} 
        \caption{Ablation study on different components in the \textit{CMU-Syn} (2s/2s) setting. \textbf{LTFU}: Local Temporal Feature Update; \textbf{GTFU}: Global Temporal Feature Update; \textbf{LRwGR}: Local Refinement with Global Representation; \textbf{GRwLR}: Global Refinement with Local Representation; \textbf{DE}: Inter-person Distance Embedding. The number of parameters (in Millions) is shown in parentheses.}
    \label{tab:Key Components}
\end{table}

\begin{figure}[ht]
\begin{center}
\includegraphics[width=0.96\linewidth]{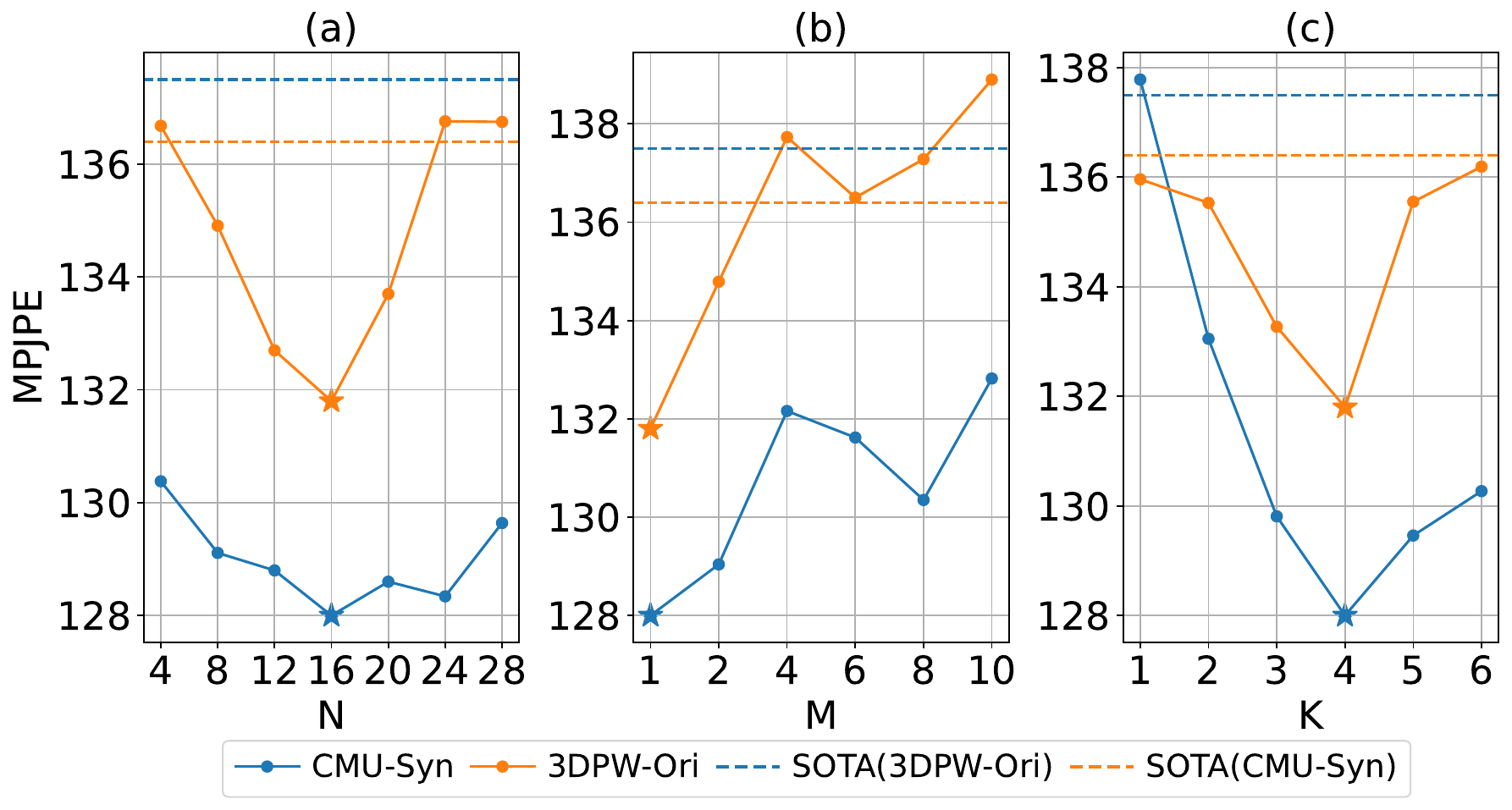}
\end{center}
\setlength{\abovecaptionskip}{-5pt} 
\setlength{\belowcaptionskip}{5pt} 
\caption{Performance curves with varied hyper-parameters in \textit{CMU-Syn} (2s/2s) and \textit{3DPW-Ori} (1030ms/900ms) settings. $N$/$M$ represent the iteration number of local/global temporal feature updates per ME block, and $K$ denotes the number of ESTFL stages.}
\label{fig:hyper}
\end{figure}

\noindent\textbf{Effectiveness of Distance Embedding.}
In the CI block, we explicitly integrate inter-person distance embedding (DE) to refine local representations by capturing spatial relationships between individuals. As illustrated in the last two rows of Tab.~\ref{tab:Key Components}, the model without the distance embedding operation exhibits decreased performance compared to our full model,  validating the effectiveness of this component.

\begin{table}[ht]
    \centering
    \large
    \renewcommand{\arraystretch}{1.3}
    \setlength{\tabcolsep}{4pt}
    \resizebox{1\columnwidth}{!}{
    \begin{tabular}{cc|ccc}
        \toprule
        Learn Multi-level feat. & Learn Cross-level feat. & MPJPE & Params & FLOPs \\
        \midrule
        Self-Attention & \textbf{CI block} & 131.7 & 3.28M & 39.83G \\
        \textbf{ME block} & Cross-Attention & 133.6 & 0.26M & 3.07G \\
        \textbf{ME block}  & \textbf{CI block} & \textbf{128.0} & \textbf{0.17M} & \textbf{1.67G} \\
        \bottomrule
    \end{tabular}
    }
    \setlength{\abovecaptionskip}{5pt} 
    \setlength{\belowcaptionskip}{-5pt} 
    \caption{Comparisons with Attention-based design in \textit{CMU-Syn (2s/2s) setting}. Network with our designs achieves superior performance with fewer parameters and lower computational cost.} 
    \label{tab:architecture_comparison}
\end{table}

\noindent\textbf{Comparison with Attention-based Design.} 
In our model, multi-level temporal features are learned through linear layers and layer normalization within the ME block, whereas cross-level interactions are achieved through learned affine transformations in the CI block. By contrast, previous models have predominantly employed Attention-based mechanisms for both feature learning and interaction. Besides comparing our method against traditional Attention-based architectures in Sect.~\ref{sec5:Experimental Results}, we further investigate the effectiveness of our ME/CI block by replacing them with Attention-based designs (details in supplementary material), as shown in Tab.~\ref{tab:architecture_comparison}. Results indicate that  Attention-based design incurs higher computational costs and yields inferior MPJPE scores. Conversely, our proposed design achieves superior performance while substantially reducing the number of parameters and computational FLOPs.

\noindent\textbf{Impact of Hyper-parameters.} 
We examine the impact of essential hyper-parameters, including the number of local/global temporal feature update iterations ($N$/$M$) and the number of ESTFL stages ($K$). As illustrated in Fig.~\ref{fig:hyper}, the optimal configuration ($N=16, M=1, K=4$) achieves the best performance for both the \textit{CMU-Syn} (2s/2s) and \textit{3DPW-Ori} (1030ms/900ms) settings. Our EMPMP model  outperforms state-of-the-art methods across most hyper-parameter variations, highlighting its robustness for hyper-parameters. Notably, our network with $M\le2$ presents consistently superior performance than cases of $M>2$, indicating that excessive global iterations may lead to over-smoothing of temporal features and thus degrade prediction accuracy.  Further ablation studies with different values of  combination parameter $\alpha$ in Eq.~(\ref{eq:combine}) are detailed in supplementary material, with $\alpha=0.2$ producing the best results.

\subsection{Qualitative Results}
\label{subsec5.6:Qualitative }
In this section, we provide a qualitative comparison between our model and existing methods using the \textit{3DPW-RC} setting, as illustrated in Fig.~\ref{fig:visualization}. The ground truth consists of two subjects performing distinct actions: the left subject remains largely static with minimal leg movement, while the right subject executes a full turn, exhibiting more dynamic motion.
Our model outperforms previous methods in predicting the movements of both subjects. For the left subject, our model  preserves subtle details of the pose with greater precision. For the right subject,  our model captures the dynamic motion while ensuring stability, maintaining temporal coherence and realism throughout the movement. Additional visualization results demonstrating the effectiveness of our model are provided in supplementary material, further highlighting the superiority of our EMPMP network.

\section{Conclusion}
\label{sec6:Conclusion}
In this paper, we introduce EMPMP, a highly effective and efficient architecture for 3D multi-person motion prediction. Our design incorporates a dual-branch  to capture multi-level spatial-temporal features and a novel cross-level interaction strategy to enable information exchange, achieving significant reductions in parameter count and computational complexity. Extensive experiments across multiple datasets validate the efficiency and accuracy of our approach, demonstrating strong predictive performance with minimal computational overhead.
 Future directions include stochastic motion prediction to capture inherent uncertainty in human movement and advanced feature fusion strategies to further enhance accuracy and robustness.


{
    \small
    \bibliographystyle{ieeenat_fullname}
    \bibliography{main}
}

\clearpage
\setcounter{page}{1}
\maketitlesupplementary
\setlength{\parindent}{0pt}

\setcounter{equation}{0}  
\renewcommand{\theequation}{\Alph{section}.\arabic{equation}}
\renewcommand{\thefigure}{\Alph{section}.\arabic{figure}}
\renewcommand{\thetable}{\Alph{section}.\arabic{table}}
\setcounter{figure}{0}  
\pretocmd{\section}{\setcounter{table}{0}}{}{}
\pretocmd{\section}{\setcounter{figure}{0}}{}{}
\pretocmd{\section}{\setcounter{equation}{0}}{}{}

\setcounter{table}{0}  

\renewcommand{\thesection}{\Alph{section}}
\setcounter{section}{0}  


\section{Additional Details on Dataset and Metric}
 
\subsection{Dataset Sources}
All datasets utilized in this study are sourced from publicly available open-source repositories, including the CMU-Mocap, MuPoTS-3D, and 3DPW datasets. For the 3DPW dataset, we adhere to the protocol outlined in the SoMoF Benchmark~\cite{adeli2021tripod,adeli2020socially}, which has been used in previous studies~\cite{vendrow2022somoformer,xu2023joint}. For the 3DPW-RC dataset, we apply the same scripts as in~\cite{xu2023joint} to remove camera movement, thereby enhancing the realism of human motion. For the CMU-Mocap and MuPoTS-3D datasets, we synthesize data based on the original datasets, following  approach in~\cite{wang2021multi}.   Since synthesized datasets from prior works are not publicly available, all experiments were conducted on our synthesized versions by running official codes of compared methods.

\subsection{Metric Formulations}
Given the predicted motion $Y=\{y_t^{p,j}\}\in \mathbb{R}^{3J\times P \times T'}$  for $P$ persons across $T'$ time frames with $J$ joints per person, along with the corresponding ground truth $\hat Y=\{\hat y_t^{p,j}\}\in \mathbb{R}^{3J\times P \times T'}$, the following metrics are used for evaluation.

\noindent\textbf{MPJPE.} 
The Mean Per Joint Position Error (MPJPE) measures the overall joint prediction accuracy by averaging errors across all time frames:
\setcounter{equation}{0}
\begin{equation}
\textbf{MPJPE} = \frac{1}{P \cdot T' \cdot J} \sum_{p=1}^{P} \sum_{t=1}^{T'} \sum_{j=1}^{J} || {y}_t^{p,j} - \hat{{y}}_t^{p,j} ||_2.
\end{equation}

\noindent\textbf{VIM.} 
The Visibility-Ignored Metric (VIM) focuses on the average joint error for a specific time frame, and the VIM score at timestep $t$ is given by:
\begin{equation}
\textbf{VIM@t} = \frac{1}{P } \sum_{p=1}^{P}  \sqrt{\sum_{j=1}^{J}  ({y}_t^{p,j} - \hat{{y}}_t^{p,j} )^2}.
\end{equation}

\noindent\textbf{JPE.} 
The Joint Precision Error (JPE) assesses both global and local joint predictions using the mean $L_2$ distance of all joints for timestep $t$:
\begin{equation}
\textbf{JPE@t} = \frac{1}{P \cdot J} \sum_{p=1}^{P} \sum_{j=1}^{J} || \mathbf{y}_t^{p,j} - \hat{\mathbf{y}}_t^{p,j} ||_2.
\end{equation}

\noindent\textbf{APE.} 
The Aligned Mean Per Joint Position Error (APE) evaluates the forecasted local motion by computing the $L_2$ distance for each joint, averaged across all joints at a given timestep $t$, with global displacement removed by subtracting the hip joint:
\begin{equation}
\textbf{APE@t} = \frac{1}{P \cdot J} \sum_{p=1}^{P} \sum_{j=1}^{J} || ({y}_t^{p,j}-{y}_{t,\textbf{hip}}^{p}) - (\hat{{y}}_t^{p,j}-\hat{{y}}_{t,\textbf{hip}}^{p}) ||_2.
\end{equation}

\noindent\textbf{FDE.} 
The Final Distance Error (FDE) quantifies the accuracy of the forecasted global trajectory by computing the $L_2$ distance for a specific timestep $t$:
\begin{equation}
\textbf{FDE@t} = \frac{1}{P \cdot J} \sum_{p=1}^{P} \sum_{j=1}^{J} || {y}_{t,\textbf{hip}}^{p} -\hat{{y}}_{t,\textbf{hip}}^{p} ||_2.
\end{equation}

These metrics provide a comprehensive evaluation of the accuracy for 3D  motion prediction task, capturing both local joint-wise pose errors and global trajectory deviations.

\section{Additional Network and Training Details}

\noindent\textbf{Details on PIPS and IPIPS Stages.}
In order to maintain the invariance of our model to the order of individuals, we introduce Permutation-Invariant Person Sorting (PIPS). Given the input \( X \in \mathbb{R}^{3J \times P \times T} \), we calculate the sum of distances between each individual and all others as:
\begin{equation}
d_{p_j} = \sum_{p_k \neq p_j}^{P} \| {x}_{1,\textbf{hip}}^{p_j} - {x}_{1,\textbf{hip}}^{p_k} \|_2, \quad j, k = 1, \dots, P.
\end{equation}
We then sort individuals in descending order based on these values, modifying the input \( X \). After processing the input through the model to obtain output \( Y' \), we apply Inverse PIPS to recover the original person order, resulting in the final output \( Y \in \mathbb{R}^{3J \times P \times T'} \).
These two stages ensures that the output of our model is unchanged for different orders of individuals in the same scene as the input.

\begin{figure*}[!t]
\begin{center}
\includegraphics[width=0.8\linewidth,height=2.4in]{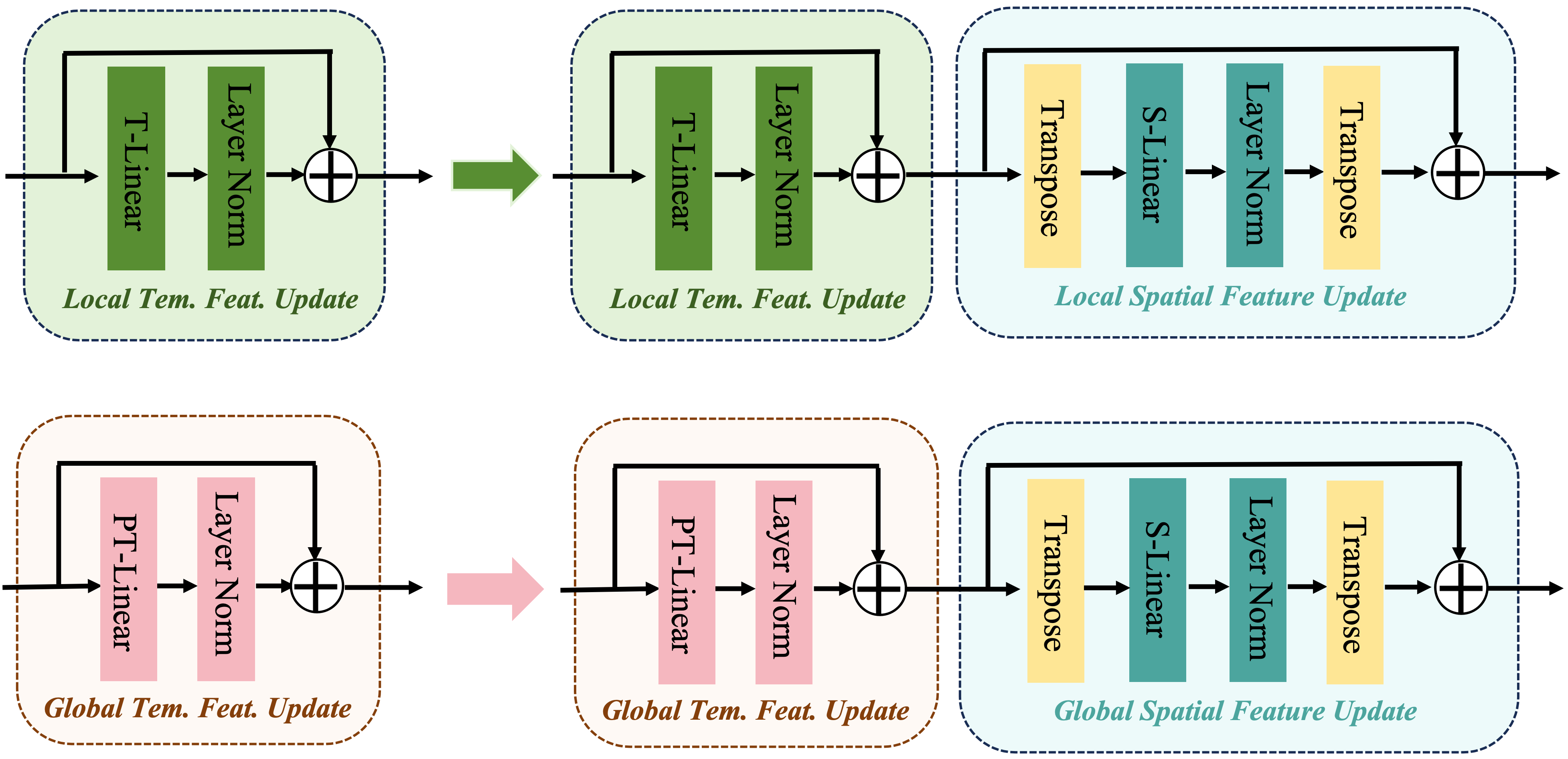}
\end{center}
\setlength{\abovecaptionskip}{-5pt} 
\setlength{\belowcaptionskip}{-10pt} 
\caption{Architectural modifications for our 0.65M-parameter EMPMP involve enhancing the Local/Global Temporal Feature Update (left) with the addition of the Local/Global Spatial Feature Update (right). These modifications update features along the spatial dimension, thereby increasing the model's expressive capacity.}
\label{fig:pre-train architecture}
\end{figure*}

\noindent\textbf{Network for Pre-training.} For the experiments in the settings of \textit{AMASS/3DPW-Ori} and \textit{AMASS/3DPW-RC}, the network is pre-trained on the AMASS dataset and fine-tuned on the 3DPW dataset. Given that the number of parameters has a significant impact on network performance, especially for lightweight architectures, we utilize a model with 0.65M parameters to increase its capacity. This is achieved by incorporating additional spatial feature updates, as illustrated in Fig.~\ref{fig:pre-train architecture}. Specifically, we introduce Local/Global Spatial Feature Update, which extend the Local/Global Temporal Feature Update. These new components maintain a similar architectural structure to the original components but operate along the spatial dimension rather than the temporal dimension. As demonstrated in Tab.~1 of the main paper, our model with 0.65M parameters achieves the best performance for the \textit{AMASS/3DPW-RC} setting and  competitive results for the \textit{AMASS/3DPW-Ori} setting. 

\noindent\textbf{Details for Pre-training Settings.}
For the experiments in the settings of \textit{AMASS/3DPW-Ori} and \textit{AMASS/3DPW-RC}, we pre-train our network on AMASS dataset for 100 epochs with an initial learning rate of $1 \times 10^{-4}$, which decays by a factor of 0.8 every 10 epochs. The strategy employed for network fine-tuning is the same as described in Sect.~3.5 of the main paper.

\noindent\textbf{Details on Attention-based Designs.}
In Sect.~4.3 of the main paper, we compared our architecture with Attention-based alternatives by replacing our ME block and CI block with two distinct Attention-based blocks: one employing Multi-Head Attention (Self-Attention) for local/global temporal feature update and the other utilizing Cross Attention for local/glocal refinement. The detailed  architectures are illustrated in Fig.~\ref{fig:Transformer Architecture}. Our results in Tab.~5 of the main paper demonstrate that our proposed model outperforms traditional Attention-based architectures, which have more parameters and require greater computational resources.

\section{More Experimental Results}
In addition to the results evaluated using MEJPE, VIM, and APE presented in the main paper, we provide additional results based on the metrics of JPE and FDE in this section. We also include the results across different key frames for VIM, JPE, APE, and FDE, along with an analysis of the computational cost.

\noindent\textbf{Results Evaluated by JPE and FDE.}
In Tab.~\ref{tab:Results on JPE and FDE}, we present the results evaluated using the JPE and FDE metrics across all the settings discussed in the main paper, along with comparisons to previous works. Our EMPMP model achieves the best JPE results, demonstrating significant superiority over the other methods. It also performs exceptionally well in FDE, ranking first in seven out of eight evaluations. Notably, the T2P model~\cite{jeong2024multi} predicts multiple trajectories ($F$) and for a fair comparison, we compute the FDE with $F=1$, while using $F=3$ for the other metrics.

\noindent\textbf{Detailed Results Across Key Frames.}
In the main paper, for all models evaluated using VIM, JPE, APE, and FDE, we report their average results across key frames. Additionally, we provide the detailed results for each individual frame, following the frame selection scheme shown in Tab.~\ref{tab:selection}. For the VIM metric on the 3DPW dataset, we adopt the same frame selection scheme as used in~\cite{xu2023joint,vendrow2022somoformer} to ensure consistency and fair comparisons. For other datasets and metrics, frames are selected at reasonable intervals to ensure comprehensive evaluation. Tab.~\ref{tab:vim_res} presents the detailed VIM results across multiple settings of the 3DPW dataset, while Tab.~\ref{tab:JPE,APE,FDE} reports the corresponding JPE, APE, and FDE results for the same dataset. Similarly, Tab.~\ref{tab:vim_res on the CMU-Syn and MuPoTS} shows the VIM results for various settings of the CMU-Syn and MuPoTS-3D datasets, and Tab.~\ref{tab:JPE,APE,FDE on cmu and MuPoTS} provides the JPE, APE, and FDE results for these datasets. These tables provide detailed results for specific key frames and their average values, where our model  achieves dominant superior performance in most comparisons, proving its effectiveness through a comprehensive evaluation of performance metrics across different timesteps.



\begin{table}[ht]
    \centering
    \small
    \renewcommand{\arraystretch}{1.0} 
    \resizebox{\columnwidth}{!}{ 
    \begin{tabular}{c|c|c|c}
        \toprule
        \midrule     
        Datasets    & 3DPW             & \multicolumn{2}{c}{CMU-Syn \& MuPoTS }     \\ \cmidrule{2-4}         
        Out Length  & 14frames (900ms) & 30frames (2s)       & 15frames (1s)  \\ \midrule

         \textbf{VIM}                    & 2, 4, 8, 10, 14  & 2, 6, 11, 21, 30    & 2, 4, 8, 10, 15                   \\ \cmidrule{2-4} 
         \textbf{JPE\&APE\&FDE}            & 7, 14            & 10, 20, 30          & 3, 9, 15                          \\
                               
        \midrule
        \bottomrule
    \end{tabular}
    }
    \setlength{\abovecaptionskip}{8pt} 
    \setlength{\belowcaptionskip}{-5pt} 
    \caption{Frame selection scheme for different datasets. }
    \label{tab:selection}
\end{table}


\noindent\textbf{Comparison on Computational Cost.}
In Tab.~\ref{tab:Computational cost}, we present the detailed computational costs of our EMPMP model and the compared methods, including GPU memory usage, computational FLOPs, and the number of parameters.  Our model demonstrates superior performance across various settings while maintaining a significantly lower number of parameters and FLOPs.

\begin{table}[ht]
    \centering
    \small
    \renewcommand{\arraystretch}{1.0} 
    \resizebox{\columnwidth}{!}{ 
    \begin{tabular}{l|>{\centering\arraybackslash}p{1.4cm}>{\centering\arraybackslash}p{1.4cm}>{\centering\arraybackslash}p{1.4cm}}
        \toprule
        \midrule     
        \multirow{2}{*}{Metrics}    & Memory   &  FLOPs        & Params      \\ 
          & (MB)    &  (G)  & (M)   \\ \midrule

        MRT~\cite{wang2021multi} {\scriptsize{$^{\rm \textcolor{blue}{'2021}}$}}                      &\textbf{2281}	&27.55	&6.29       \\
        SoMoFormer~\cite{vendrow2022somoformer} {\scriptsize{$^{\rm \textcolor{blue}{'2022}}$}}       &6308	&113.37	&12.91                 \\
        TBIFormer~\cite{peng2023trajectory} {\scriptsize{$^{\rm \textcolor{blue}{'2023}}$}}       &2826	&15.64	&7.31                \\
        JRT~\cite{xu2023joint} {\scriptsize{$^{\rm \textcolor{blue}{'2023}}$}}            &15544 &767.74	&3.68                \\
        T2P~\cite{jeong2024multi} {\scriptsize{$^{\rm \textcolor{blue}{'2024}}$}}                        &4304	&51.67	&4.60                \\
        Ours           &2674	&\textbf{1.67}	&\textbf{0.17}                \\
        \midrule
        \bottomrule
    \end{tabular}
    }
    \setlength{\abovecaptionskip}{8pt} 
    \setlength{\belowcaptionskip}{-5pt} 
    \caption{Comparisons of computational cost for different models in the \textit{CMU-Syn} (2s/2s) setting.}
    \label{tab:Computational cost}
\end{table}

\section{Additional Ablation Study}

\noindent\textbf{Effectiveness of Learned Affine Transformations.}
In the CI block, we incorporate both scale and translation for local representation refinement, while employing translation alone for global representation refinement. To assess the effectiveness of this design, we conducted an ablation study exploring various transformation choices for local/global representation refinement. As presented in Tab.~\ref{tab:affine}, the model that utilizes scale and translation for local refinement, and translation for global refinement, yields the best performance compared to alternative configurations.

\begin{table}[ht]
    \centering
    \small
    \resizebox{1\linewidth}{!}{ 
    \begin{tabular}{>{\centering\arraybackslash}p{1.2cm} >{\centering\arraybackslash}p{1.2cm} >{\centering\arraybackslash}p{1.2cm} >{\centering\arraybackslash}p{1.2cm} | >{\centering\arraybackslash}p{1.2cm}}
        \toprule
        \midrule
        \multicolumn{2}{c|}{Local Refinement} &\multicolumn{2}{c|}{Global Refinement}& \multirow{2}{*}[-2pt]{\textbf{MPJPE}} \\ \cmidrule{1-4} 
        \multicolumn{1}{c|}{\textbf{Scale}} & \multicolumn{1}{c|}{\textbf{Translation}}& \multicolumn{1}{c|}{\textbf{Scale}} & \multicolumn{1}{c|}{\textbf{Translation}}  \\ \midrule
        × & \checkmark & \checkmark & \checkmark & 129.9  \\ 
        \checkmark & × & \checkmark & \checkmark & 129.4  \\ 
        \checkmark & \checkmark & × & \checkmark & \textbf{128.0}  \\ 
        \checkmark & \checkmark & \checkmark & × & 131.3  \\ 
        \checkmark & \checkmark & \checkmark & \checkmark & 129.0 \\ \midrule
        \bottomrule
    \end{tabular}
    }
    \setlength{\abovecaptionskip}{8pt} 
    \setlength{\belowcaptionskip}{-5pt} 
    \caption{Ablation study on learned affine transformations for local and global representation refinement. }
    \label{tab:affine}
\end{table}

\noindent\textbf{Ablation Study on Loss Function.}
Our loss function comprises two components: mean joint loss and velocity loss. As demonstrated in Tab.~\ref{tab:abalation study in loss function}, the network that excludes velocity loss performs worse in the \textit{3DPW-Ori}, \textit{3DPW-RC}, and \textit{CMU-Syn} settings, thereby highlighting the necessity of incorporating velocity loss.

\begin{table}[ht]
    \centering
    \renewcommand{\arraystretch}{1.0} 
    \resizebox{\columnwidth}{!}{ 
    \begin{tabular}{l|>{\centering\arraybackslash}p{1.4cm}|>{\centering\arraybackslash}p{1.4cm}|>{\centering\arraybackslash}p{1.4cm}|>{\centering\arraybackslash}p{1.4cm}}
        \toprule
        \midrule     
        \multirow{2}{*}{Settings}    & 3DPW-Ori   &  3DPW-RC         & \multicolumn{2}{c}{\multirow{2}{*}{CMU-Syn}}       \\ \cmidrule{2-5}
        In/Out Length  & \multicolumn{2}{c|}{1030ms/900ms}      & 2s/2s    &1s/1s  \\ \midrule

         {EMPMP-w/o-$\mathcal{L}^{Vel}$}                    &137.7&103.7	&130.0& 74.7              \\
         {EMPMP}            &\textbf{131.8}&\textbf{99.5}	&\textbf{128.0} &\textbf{73.5}                    \\
                               
        \midrule
        \bottomrule
    \end{tabular}
    }
    \setlength{\abovecaptionskip}{8pt} 
    \setlength{\belowcaptionskip}{-5pt} 
    \caption{Ablation study on velocity loss in \textit{3DPW-Ori}, \textit{3DPW-RC} and \textit{CMU-Syn} settings, evaluated using MPJPE.}
    \label{tab:abalation study in loss function}
\end{table}

\noindent\textbf{Ablation Study on Combination Parameter.}
The combination parameter $\alpha$ in Eq.~(11) of the main paper controls the balance between global and local representations. We present the results of our EMPMP model with varying values of  $\alpha$ in the \textit{3DPW-RC} setting. As shown in Tab.~\ref{tab:abalation study in combination para}, the optimal results for MPJPE, VIM, and APE were achieved with $\alpha = 0.2$.

\begin{table}[ht]
    \centering
    \Large
    \renewcommand{\arraystretch}{1.1} 
    \resizebox{\columnwidth}{!}{ 
    \begin{tabular}{c|ccccccccc}
        \toprule
        \midrule     
        $\alpha$    & 0  &  0.1         & 0.2 &0.3& 0.4 &0.6 &0.8 &1.0       \\ \cmidrule{1-9}        
         
         {\textbf{MPJPE}}          &100.6	&102.9	&\textbf{99.5}	&99.8	&102.1  &100.3	&100.4	&101.0	            \\
         {\textbf{VIM}}            &41.7	&42.7	&\textbf{41.3}&	41.5	&42.3  &41.5	&41.8	&42.6	                  \\
         {\textbf{APE}}          & 97.3	&99.2	&\textbf{96.6}	&99.4	&96.9 &97.0	&97.4	&97.9	                 \\          
        \midrule
        \bottomrule
    \end{tabular}
    }
    \setlength{\abovecaptionskip}{8pt} 
    \setlength{\belowcaptionskip}{-5pt} 
    \caption{Ablation study of the combination parameter $\alpha$ in the \textit{3DPW-RC} setting.}
    \label{tab:abalation study in combination para}
\end{table}



\section{More Qualitative Results}
In this section, we provide additional visualization results. Fig.~\ref{fig:mocap15to15} presents qualitative results in the \textit{CMU-Syn} (1s/1s) setting, while Fig.~\ref{fig:mupots15to15} showcases results in the \textit{CMU-Syn/MuPoTS} (1s/1s) setting. These results highlight the effectiveness of our EMPMP model, as it generates motion sequences that exhibit a closer alignment with the ground truth, thereby preserving realistic motion dynamics.

\begin{table*}[ht]
    \centering
    \small
    \renewcommand{\thetable}{C.3} 
    \renewcommand{\arraystretch}{1.0} 
    \resizebox{\textwidth}{!}{ 
    \begin{tabular}{>{\centering\arraybackslash}p{0.5cm}l|>{\centering\arraybackslash}p{1.4cm}|>{\centering\arraybackslash}p{1.4cm}|>{\centering\arraybackslash}p{1.2cm}|>{\centering\arraybackslash}p{1.2cm}|>{\centering\arraybackslash}p{1.4cm}|>{\centering\arraybackslash}p{1.4cm}|>{\centering\arraybackslash}p{1.2cm}|>{\centering\arraybackslash}p{1.2cm}}
        \toprule
        \midrule
        \multirow{3}{*}{\rotatebox{90}{Metric}}
        &\multicolumn{1}{l|}{\multirow{2}{*}{Settings}}         & \multirow{2}{*}{3DPW-Ori} &\multirow{2}{*}{3DPW-RC}  & \multicolumn{2}{c|}{\multirow{2}{*}{CMU-Syn}} &{AMASS\slash3DPW-Ori} &AMASS\slash3DPW-RC & \multicolumn{2}{c}{\multirow{2}{*}{CMU-Syn/MuPoTS}}   \\ \cmidrule{3-10}
        &\multicolumn{1}{l|}{In/Out Length}   &\multicolumn{2}{c|}{1030ms / 900ms}   & \multicolumn{1}{c|}{2s / 2s} & 1s / 1s &\multicolumn{2}{c|}{1030ms / 900ms} & \multicolumn{1}{c|}{2s / 2s} & 1s / 1s              \\ \midrule
        \multirow{6}{*}{\rotatebox{90}{\textbf{JPE}}} 
                & MRT~\cite{wang2021multi} {\scriptsize{$^{\rm \textcolor{blue}{'2021}}$}} & 236.4	&182.0	&197.0	&92.2	&208.8	&169.4	&228.3	&106.6                           \\
                               & SoMoFormer~\cite{vendrow2022somoformer} {\scriptsize{$^{\rm \textcolor{blue}{'2022}}$}}    &207.0	&143.7	&184.1	&86.2	&167.9	&130.3	&200.5	&98.5 \\
                               & TBIFormer~\cite{peng2023trajectory} {\scriptsize{$^{\rm \textcolor{blue}{'2023}}$}}       &202.8	&163.2	&214.8	&103.4	&197.1	&149.6	&218.0	&111.4                       \\
                               & JRT~\cite{xu2023joint} {\scriptsize{$^{\rm \textcolor{blue}{'2023}}$}}   &223.8	&154.3	&193.7	&94.2	&196.5	&138.1	&199.1	&102.9                        \\ 
                               & T2P~\cite{jeong2024multi} {\scriptsize{$^{\rm \textcolor{blue}{'2024}}$}}     &190.9	&150.4	&167.5	&84.6	&173.4	&128.8	&199.1	&109.7       \\ 
        \rowcolor{gray!20}     & Ours               &\textbf{184.8}&\textbf{128.1}&\textbf{155.2}&\textbf{79.5}&\textbf{164.1}&\textbf{118.6}&\textbf{193.7}&\textbf{98.3}    \\ \midrule
        
        \multirow{6}{*}{\rotatebox{90}{\textbf{FDE}}} 
        & MRT~\cite{wang2021multi} {\scriptsize{$^{\rm \textcolor{blue}{'2021}}$}}           &192.8	&136.6	&168.9	&69.0	&167.2	&127.5	&178.4	&77.2                                    \\
                               & SoMoFormer~\cite{vendrow2022somoformer} {\scriptsize{$^{\rm \textcolor{blue}{'2022}}$}}        & 166.0	 &95.3	 &157.7	 &63.6	 &{133.0}	 &87.2	 &158.8	 &72.6                              \\
                               & TBIFormer~\cite{peng2023trajectory} {\scriptsize{$^{\rm \textcolor{blue}{'2023}}$}}       &156.9	&117.8	&184.3&79.3	&154.5	&108.4	&168.5	&78.7        \\
                               & JRT~\cite{xu2023joint} {\scriptsize{$^{\rm \textcolor{blue}{'2023}}$}}             &181.3	&103.3	&162.8	&70.5	&158.0	&88.5	&\textbf{144.7}	&75.8                 \\ 
                                & T2P{\scriptsize{$^{\rm \textcolor{blue}{F=1}}$}}~\cite{jeong2024multi}  {\scriptsize{$^{\rm \textcolor{blue}{'2024}}$}}     &165.3	&106.9	&169.8	&73.4	&146.6	&93.7	&164.4	&78.6             \\
        \rowcolor{gray!20}     & Ours               &\textbf{148.8}&\textbf{86.9}&\textbf{128.1}&\textbf{57.9}&\textbf{132.2}&\textbf{79.8}&150.6&\textbf{70.6}     \\ \midrule
        \bottomrule
    \end{tabular}
    }
    \setlength{\abovecaptionskip}{8pt} 
    \setlength{\belowcaptionskip}{-5pt} 
    \caption{Results in multiple settings for JPE and FDE. Our EMPMP network achieves the best performance in most comparisons.}
    \label{tab:Results on JPE and FDE}
\end{table*}

\begin{table*}[!t]
\small
\renewcommand{\thetable}{C.4} 
\begin{center}
    \begin{tabular}{l|cccccc|cccccc}
    \toprule
    \toprule
    \multirow{1}{*}{Settings} & \multicolumn{6}{c}{3DPW-Ori}                & \multicolumn{6}{|c}{3DPW-RC}           \\
    \cmidrule{2-13}
    \multirow{1}{*}{Selected Frames}       & 2 & 4 & 8 & 10 & 14 & AVG   & 2 & 4 & 8 & 10 & 14& AVG \\
    \midrule 
    MRT~\cite{wang2021multi} {\scriptsize{$^{\rm \textcolor{blue}{'2021}}$}}  
    &19.6	&36.5	&68.9	&86.4	&123.1	&66.9  &18.5	&33.8	&59.3	&71.5	&93.2	&55.2 \\  
    
    SoMoFormer~\cite{vendrow2022somoformer} {\scriptsize{$^{\rm \textcolor{blue}{'2022}}$}}  
    &13.0	&28.5	&59.5	&76.4	&111.7	&57.8  &12.3	&26.5	&49.9	&59.5	&74.0	&44.4 \\  
    
    TBIFormer~\cite{peng2023trajectory} {\scriptsize{$^{\rm \textcolor{blue}{'2023}}$}}  
    &17.4	&33.5	&63.4	&78.2	&108.5	&60.2  &15.7	&30.3	&56.1	&67.6	&86.7	&51.2  \\  
    
    JRT~\cite{xu2023joint} {\scriptsize{$^{\rm \textcolor{blue}{'2023}}$}}  
    &12.5	&29.0	&61.6	&78.1	&111.5	&58.5  &12.6	&28.7	&53.6	&63.1	&77.1	&47.0 \\  
    
    T2P~\cite{jeong2024multi} {\scriptsize{$^{\rm \textcolor{blue}{'2024}}$}}  
    &16.6	&31.6	&59.6	&73.0	&\textbf{100.7}	&56.3  &15.2	&29.3	&51.5	&60.9	&77.8	&46.9 \\  

    \rowcolor{gray!20} \textbf{Ours}    &\textbf{12.3}	 &\textbf{26.2} 	 &\textbf{55.1}	 &\textbf{70.5}	 &102.6	 &\textbf{53.3}  &\textbf{11.7}	&\textbf{24.5}	&\textbf{46.3}	&\textbf{55.2}	&\textbf{69.1}	&\textbf{41.3}  \\
    \midrule 
    
    \multirow{1}{*}{Settings} & \multicolumn{6}{c}{AMASS/3DPW-Ori}                & \multicolumn{6}{|c}{AMASS/3DPW-RC}           \\
    \cmidrule{2-13}
    \multirow{1}{*}{Selected Frames}       & 2 & 4 & 8 & 10 & 14 & AVG   & 2 & 4 & 8 & 10 & 14& AVG \\
    \midrule
    MRT~\cite{wang2021multi} {\scriptsize{$^{\rm \textcolor{blue}{'2021}}$}}  
    &21.8	&39.1	&65.1	&75.9	&94.1	&59.2  &20.8	&36.4	&58.2	&66.6	&79.4	&52.3 \\  
    
    SoMoFormer~\cite{vendrow2022somoformer} {\scriptsize{$^{\rm \textcolor{blue}{'2022}}$}}  
    &\textbf{9.1}	&\textbf{21.3}	&\textbf{47.5}	&\textbf{61.6}	&\textbf{91.9}	&\textbf{46.3}  &10.6 &22.8 &44.5 &54.0 &68.4 &40.0 \\  
    
    TBIFormer~\cite{peng2023trajectory} {\scriptsize{$^{\rm \textcolor{blue}{'2023}}$}}  
    &13.3	&28.4	&58.9	&74.7	&106.8	&56.4  &13.7	&27.0	&52.1	&62.8	&81.4	&47.4  \\  
    
    JRT~\cite{xu2023joint} {\scriptsize{$^{\rm \textcolor{blue}{'2023}}$}}  
    &{9.5}	&22.1	&48.7	&62.8	&92.8	&47.2  &\textbf{9.5}	&21.7	&44.1	&53.4	&68.8	&39.5 \\  
    
    T2P~\cite{jeong2024multi} {\scriptsize{$^{\rm \textcolor{blue}{'2024}}$}}  
    &11.0	&23.2	&50.8	&65.7	&96.3	&49.4  &12.0	&24.3	&46.4	&58.1	&71.2	&42.4 \\  
    
    \rowcolor{gray!20} \textbf{Ours}   &10.5	&23.1	&49.8	&64.6	&95.2	&48.6 &9.8	&\textbf{21.6}	&\textbf{42.6}	&\textbf{51.8}	&\textbf{66.2}	&\textbf{38.4} \\
    \bottomrule
    \bottomrule
    \end{tabular}
    \setlength{\abovecaptionskip}{8pt} 
    \setlength{\belowcaptionskip}{-5pt} 
\caption{Detailed VIM results on the 3DPW dataset across different settings are presented. Our EMPMP demonstrates the best performance in the majority of comparisons across three out of the four settings.}
\label{tab:vim_res}
\end{center}
\end{table*}

\begin{table*}[!t]
\begin{center}
\small
\renewcommand{\thetable}{C.5} 
\resizebox{\textwidth}{!}{ 
    \begin{tabular}{cl|ccc|ccc|ccc|ccc}
    \toprule
    \toprule
    \multirow{2}{*}{\rotatebox{90}{Metric}}&
    \multirow{1}{*}{Settings} & \multicolumn{3}{c|}{3DPW-Ori}   & \multicolumn{3}{c|}{3DPW-RC}  & \multicolumn{3}{c}{AMASS/3DPW-Ori}   & \multicolumn{3}{|c}{AMASS/3DPW-RC}      \\
    \cmidrule{3-14}
    &\multirow{1}{*}{Selected Frames}       & 7 & 14 & AVG & 7 & 14 & AVG   & 7 & 14 & AVG & 7 & 14& AVG \\
    \midrule 
    
    \multirow{6}{*}{\rotatebox{90}{\textbf{JPE}}} 
    &MRT~\cite{wang2021multi} {\scriptsize{$^{\rm \textcolor{blue}{'2021}}$}}  
    &150.0	&322.9	&236.4  &128.1	&235.9	&182.0  &133.5	&284.2	&208.8  &121.7	&217.1	&169.4 \\  
    
    &SoMoFormer~\cite{vendrow2022somoformer} {\scriptsize{$^{\rm \textcolor{blue}{'2022}}$}}  
    &125.3	&288.8	&207.0  &105.0	&182.5	&143.7  &101.0	&{234.9}	&{167.9}  &92.3	&168.4	&130.3 \\  
    
    &TBIFormer~\cite{peng2023trajectory} {\scriptsize{$^{\rm \textcolor{blue}{'2023}}$}}  
    &132.1	&273.5	&202.8  &116.1	&210.3	&163.2  &124.2	&270.0	&197.1  &105.3	&193.9	&149.6 \\  
    
    &JRT~\cite{xu2023joint} {\scriptsize{$^{\rm \textcolor{blue}{'2023}}$}}  
    &138.7	&308.9	&223.8  &116.6	&192.0	&154.3  &116.3	&276.7	&196.5  &99.7	&176.5	&138.1 \\  
    
    &T2P~\cite{jeong2024multi} {\scriptsize{$^{\rm \textcolor{blue}{'2024}}$}}  
    &126.9	&\textbf{255.0}	&190.9  &110.2	&190.6	&150.4  &111.1	&235.7	&173.4  &90.2	&167.5	&128.8 \\  
    
    \rowcolor{gray!20}  
    &\textbf{Ours}  
    &\textbf{110.9}	&258.7	&\textbf{184.8}  &\textbf{95.5}	&\textbf{160.7}	&\textbf{128.1}  &\textbf{98.5}	&\textbf{229.7}	&\textbf{164.1}  &\textbf{85.3}	&\textbf{151.9}	&\textbf{118.6} \\  
    \midrule
    
    \multirow{6}{*}{\rotatebox{90}{\textbf{APE}}} 
    &MRT~\cite{wang2021multi} {\scriptsize{$^{\rm \textcolor{blue}{'2021}}$}}  
    &103.3	&146.9	&125.1  &102.3	&145.0	&123.6  &95.1	&135.6	&115.3 &90.9	&130.7	&110.8   \\  
    
    &SoMoFormer~\cite{vendrow2022somoformer} {\scriptsize{$^{\rm \textcolor{blue}{'2022}}$}}  
    &92.0	&144.7	&118.3  &91.8	&138.0	&114.9& 74.9	&120.2	&97.5   &78.4	&124.4	&101.4 \\  
    
    &TBIFormer~\cite{peng2023trajectory} {\scriptsize{$^{\rm \textcolor{blue}{'2023}}$}}  
    &94.9	&137.0	&115.9  &94.3	&136.7	&115.5  &87.9	&132.8	&110.3  &85.3	&130.8	&108.0 \\  
    
    &JRT~\cite{xu2023joint} {\scriptsize{$^{\rm \textcolor{blue}{'2023}}$}}  
    &99.0	&147.0	&123.0  &97.6	&143.5	&120.5  &87.0	&141.3	&114.1  &85.4	&139.7	&112.5 \\  
    
    &T2P~\cite{jeong2024multi} {\scriptsize{$^{\rm \textcolor{blue}{'2024}}$}}  
    &92.0	&138.3	&115.1  &92.0	&138.3	&115.1  &83.1	&137.1	&110.1  &82.1	&135.3	&108.7 \\  

    \rowcolor{gray!20} 
    & \textbf{Ours}    
    &\textbf{78.0} &\textbf{119.3} &\textbf{98.6} &\textbf{75.4} &\textbf{117.9} &\textbf{96.6} 
    &\textbf{73.8} &\textbf{116.2} &\textbf{95.0} &\textbf{70.9} &\textbf{110.4} &\textbf{90.6}  \\
    \midrule
    
    \multirow{6}{*}{\rotatebox{90}{\textbf{FDE}}} 
    &MRT~\cite{wang2021multi} {\scriptsize{$^{\rm \textcolor{blue}{'2021}}$}}  
    &105.7	&280.0	&192.8  &87.6	&185.7	&136.6  &90.9	&243.6	&167.2  &82.4	&172.7	&127.5 \\  
    
    &SoMoFormer~\cite{vendrow2022somoformer} {\scriptsize{$^{\rm \textcolor{blue}{'2022}}$}}  
    &86.9	&245.2	&166.0  &63.4	&127.3	&95.3  &70.9	&195.2	&133.0  &59.4	&115.1	&87.2 \\  
    
    &TBIFormer~\cite{peng2023trajectory} {\scriptsize{$^{\rm \textcolor{blue}{'2023}}$}}  
    &89.5	&224.4	&156.9  &74.7	&160.9	&117.8  &87.1	&221.9	&154.5  &70.4	&146.4	&108.4 \\  
    
    &JRT~\cite{xu2023joint} {\scriptsize{$^{\rm \textcolor{blue}{'2023}}$}}  
    &99.7	&263.0	&181.3  &74.8	&131.9	&103.3  &83.1	&233.0	&158.0  &60.6	&116.4	&88.5 \\  
    
    &T2P{\scriptsize{$^{\rm \textcolor{blue}{F=1}}$}}~\cite{jeong2024multi} {\scriptsize{$^{\rm \textcolor{blue}{'2024}}$}}  
    &84.5	&246.1	&165.3  &68.6	&145.3	&106.9  &75.4	&217.9	&146.6  &63.7	&123.7	&93.7 \\  
    \rowcolor{gray!20} 
    & \textbf{Ours}   
    &\textbf{78.3} &\textbf{219.4} &\textbf{148.8} &\textbf{59.0} &\textbf{114.8} &\textbf{86.9} 
    &\textbf{69.6} &\textbf{194.8} &\textbf{132.2} &\textbf{54.1} &\textbf{105.6} &\textbf{79.8}  \\
    \midrule
    \bottomrule
    \end{tabular}
    }
    \setlength{\abovecaptionskip}{8pt} 
    \setlength{\belowcaptionskip}{-5pt} 
    \caption{Detailed JPE,APE, and FDE results on the 3DPW dataset across different settings, with our model demonstrating dominant superiority over the compared methods. }
    \label{tab:JPE,APE,FDE}
    \end{center}
\end{table*}

\begin{table*}[!t]
\small
\renewcommand{\thetable}{C.6} 
    \begin{center}
    \begin{tabular}{l|cccccc|cccccc}
    \toprule
    \toprule
    \multirow{1}{*}{Settings} & \multicolumn{6}{c}{CMU-Syn (2s/2s)}                & \multicolumn{6}{|c}{CMU-Syn (1s/1s)}           \\
    \cmidrule{2-13}
    \multirow{1}{*}{Selected Frames}       & 2 & 6 & 11 & 21 & 30 & AVG   & 2 & 4 & 8 & 10 & 15& AVG \\
    \midrule 
    MRT~\cite{wang2021multi} {\scriptsize{$^{\rm \textcolor{blue}{'2021}}$}} 
    &14.6	&39.3	&58.6	&87.8	&107.6	&61.5 &11.4	&22.2	&39.5	&46.4	&62.4	&36.3 \\
    
    SoMoFormer~\cite{vendrow2022somoformer} {\scriptsize{$^{\rm \textcolor{blue}{'2022}}$}} 
    &9.8	&31.9	&51.6	&84.6	&105.6	&56.7 &\textbf{8.4}	&19.0	&37.2	&44.9	&62.7	&34.4 \\
    
    TBIFormer~\cite{peng2023trajectory} {\scriptsize{$^{\rm \textcolor{blue}{'2023}}$}} 
    &14.6	&39.0	&59.9	&93.5	&116.4	&64.6 &11.9	&24.2	&43.8	&51.6	&70.4	&40.3 \\
    
    JRT~\cite{xu2023joint} {\scriptsize{$^{\rm \textcolor{blue}{'2023}}$}} 
    &\textbf{9.2}	&\textbf{29.3}	&52.7	&81.9	&109.9	&56.6 &9.3	&20.3	&37.6	&44.7	&58.4	&34.0 \\
    
    T2P~\cite{jeong2024multi} {\scriptsize{$^{\rm \textcolor{blue}{'2024}}$}} 
    &14.4	&36.4	&52.7	&75.1	&94.6	&54.6 &10.4	&21.1	&37.6	&43.5	&58.8	&34.2 \\
    
    \rowcolor{gray!20} \textbf{Ours} 
    &11.8	&32.8	&\textbf{48.4}	&\textbf{69.6}	&\textbf{88.8}	&\textbf{50.2} &8.9	&\textbf{18.8}	&\textbf{35.0}	&\textbf{41.8}	&\textbf{57.8}	&\textbf{32.4} \\  
    \midrule 
    
    \multirow{1}{*}{Settings} & \multicolumn{6}{c}{CMU-Syn/MuPoTS (2s/2s)}                & \multicolumn{6}{|c}{CMU-Syn/MuPoTS (1s/1s)}           \\
    \cmidrule{2-13}
    \multirow{1}{*}{Selected Frames}      & 2 & 6 & 11 & 21 & 30 & AVG   & 2 & 4 & 8 & 10 & 15& AVG \\
    \midrule
    MRT~\cite{wang2021multi} {\scriptsize{$^{\rm \textcolor{blue}{'2021}}$}} 
    &13.3 &35.3 &61.6 &104.2 &136.3 &70.1 &12.8 &23.7 &43.8 &53.3 &74.4 &41.6 \\

    SoMoFormer~\cite{vendrow2022somoformer} {\scriptsize{$^{\rm \textcolor{blue}{'2022}}$}} 
    &\textbf{12.3} &32.6 &\textbf{56.3} &94.2 &123.2 &63.7 &\textbf{12.1} &\textbf{22.1} &\textbf{41.3} &\textbf{50.5} &70.9 &\textbf{39.3} \\

    TBIFormer~\cite{peng2023trajectory} {\scriptsize{$^{\rm \textcolor{blue}{'2023}}$}} 
    &14.1 &37.0 &62.0 &99.1 &129.4 &68.3 &13.4 &25.1 &46.7 &56.7 &78.2 &44.0 \\

    JRT~\cite{xu2023joint} {\scriptsize{$^{\rm \textcolor{blue}{'2023}}$}} 
    &13.6 &34.5 &56.4 &93.7 &125.5 &64.7 &14.4 &26.0 &44.0 &52.6 &\textbf{69.2} &41.2 \\

    T2P~\cite{jeong2024multi} {\scriptsize{$^{\rm \textcolor{blue}{'2024}}$}} 
    &14.5 &37.6 &60.2 &\textbf{91.8} &\textbf{116.1} &64.0 &13.6 &25.0 &44.6 &53.6 &73.3 &42.0 \\

    \rowcolor{gray!20} \textbf{Ours} 
    &12.7 &\textbf{32.5} &\textbf{56.3} &92.6 &120.7 &\textbf{62.9} &12.6 &23.2 &43.2 &52.4 &71.9 &40.6 \\  
    \midrule
    \bottomrule
    \end{tabular}
    \setlength{\abovecaptionskip}{8pt} 
    \setlength{\belowcaptionskip}{-5pt} 
    \caption{Detailed VIM results on the CMU-Syn and MuPoTS-3D dataset across different settings. Our model achieves the best average results across three out of four settings.}
    \label{tab:vim_res on the CMU-Syn and MuPoTS}
    \end{center}
\end{table*}

\begin{table*}[!t]
\begin{center}
\Large
\renewcommand{\thetable}{C.7} 
\resizebox{\textwidth}{!}{ 
    \renewcommand{\arraystretch}{1.1} 
    \begin{tabular}{cl|cccc|cccc|cccc|cccc}
    \toprule
    \toprule
    \multirow{2}{*}{\rotatebox{90}{Metric}}&
    \multirow{1}{*}{Settings} & \multicolumn{4}{c|}{CMU-Syn (2s/2s)}   & \multicolumn{4}{c|}{CMU-Syn (1s/1s)}  & \multicolumn{4}{c}{CMU-Syn/MuPoTS (2s/2s)}   & \multicolumn{4}{|c}{CMU-Syn/MuPoTS (1s/1s)}      \\
    \cmidrule{3-18}
    &\multirow{1}{*}{Selected Frames}       & 10 & 20&30 & AVG & 3 & 9&15 & AVG   & 10 & 20&30 & AVG & 3 & 9&15 & AVG \\
    \midrule 
    
    \multirow{6}{*}{\rotatebox{90}{\textbf{JPE}}} 
    &MRT~\cite{wang2021multi} \raisebox{0ex}{\large{$^{\rm \textcolor{blue}{2021}}$}}
    &125.5	&203.8	&261.7	&197.0  &35.6	&95.8	&145.4	&92.2  &129.1	&231.6	&324.4	&228.3  &41.2	&108.7	&169.9	&106.6 \\
    
    &SoMoFormer~\cite{vendrow2022somoformer} \raisebox{0ex}{\large{$^{\rm \textcolor{blue}{2022}}$}}
    &105.0	&193.6	&253.6	&184.0  &\textbf{26.9}	&88.2	&143.5	&86.2  &111.8	&205.4	&284.4	&200.5  &\textbf{37.8}	&\textbf{100.6}	&157.1	&98.5 \\
    
    &TBIFormer~\cite{peng2023trajectory} \raisebox{0ex}{\large{$^{\rm \textcolor{blue}{2023}}$}}
    &132.9	&223.4	&288.2	&214.8  &38.1	&106.8	&165.3	&103.4  &126.6	&221.0	&306.5	&218.0  &42.8	&114.1	&177.5	&111.4 \\
    
    &JRT~\cite{xu2023joint} \raisebox{0ex}{\large{$^{\rm \textcolor{blue}{2023}}$}}
    &115.1	&200.6	&265.5	&193.7  &29.9	&97.0	&155.7	&94.2  &117.5	&203.0	&277.0	&199.1  &42.1	&105.1	&161.7	&102.9 \\
    
    &T2P~\cite{jeong2024multi} \raisebox{0ex}{\large{$^{\rm \textcolor{blue}{2024}}$}}  
    &113.2	&167.9	&221.6	&167.5  &32.3	&87.3	&134.2	&84.6  &127.4	&203.0	&\textbf{267.1}	&199.1  &46.6	&114.3	&168.4	&109.7 \\
    
    \rowcolor{gray!20} 
    &\textbf{Ours} 
    &\textbf{100.8}	&\textbf{154.6}	&\textbf{210.4}	&\textbf{155.2}  &27.4	&\textbf{81.5}	&\textbf{129.8}	&\textbf{79.5}  &\textbf{109.3}	&\textbf{196.3}	&275.6	&\textbf{193.7}  &38.5	&101.3	&\textbf{155.3}	&\textbf{98.3} \\
    \midrule

    \multirow{6}{*}{\rotatebox{90}{\textbf{APE}}} 
    &MRT~\cite{wang2021multi} \raisebox{0ex}{\large{$^{\rm \textcolor{blue}{2021}}$}}
    &82.5	&99.0	&105.1	&95.5  &30.5	&67.0	&82.6	&60.0  &97.1	&143.5	&164.8	&135.1  &38.8	&86.6	&118.4	&81.2 \\
    
    &SoMoFormer~\cite{vendrow2022somoformer} \raisebox{0ex}{\large{$^{\rm \textcolor{blue}{2022}}$}}
    &68.0	&91.6	&101.7	&87.1  &24.2	&62.5	&78.9	&55.2  &93.7	&137.8	&161.1	&130.8  &38.6	&86.1	&117.2	&80.6 \\
    
    &TBIFormer~\cite{peng2023trajectory} \raisebox{0ex}{\large{$^{\rm \textcolor{blue}{2023}}$}}
    &83.3	&100.1	&107.3	&96.9  &32.0	&72.1	&86.7	&63.6  &101.8	&144.6	&166.8	&137.7  &39.9	&90.1	&123.2	&84.4 \\
    
    &JRT~\cite{xu2023joint} \raisebox{0ex}{\large{$^{\rm \textcolor{blue}{2023}}$}}
    &80.5	&99.4	&107.6	&95.8  &26.3	&68.8	&89.9	&61.6  &93.1	&132.2	&\textbf{151.9}	&125.7  &37.2	&84.2	&113.9	&78.4 \\
    
    &T2P~\cite{jeong2024multi} \raisebox{0ex}{\large{$^{\rm \textcolor{blue}{2024}}$}}  
    &79.0	&98.7	&105.9	&94.5  &28.2	&67.4	&84.1	&59.9  &109.2	&155.1	&178.9	&147.7  &44.1	&99.7	&134.8	&92.8 \\
    
    \rowcolor{gray!20} 
    &\textbf{Ours}    
    &\textbf{67.6}	&\textbf{86.1}	&\textbf{95.9}	&\textbf{83.2}  &\textbf{24.0}	&\textbf{58.4}	&\textbf{74.6}	&\textbf{52.3}  &\textbf{89.0}	&\textbf{131.9}	&153.9	&\textbf{124.9}  &\textbf{37.0}	&\textbf{81.6}	&\textbf{110.0}	&\textbf{76.2} \\
    \midrule

    \multirow{6}{*}{\rotatebox{90}{\textbf{FDE}}} 
    &MRT~\cite{wang2021multi} \raisebox{0ex}{\large{$^{\rm \textcolor{blue}{2021}}$}}  
    &94.3	&177.2	&235.4	&168.9  &21.7	&69.0	&116.3	&69.0  &87.8	&177.5	&270.1	&178.4  &28.8	&77.2	&125.8	&77.2 \\
    
    &SoMoFormer~\cite{vendrow2022somoformer} \raisebox{0ex}{\large{$^{\rm \textcolor{blue}{2022}}$}} 
    &77.5	&167.5	&228.1	&157.7  &\textbf{14.2}	&60.9	&115.8	&63.6  &81.3	&159.7	&235.4	&158.8  &28.4	&73.4	&116.2	&72.6 \\
    
    &TBIFormer~\cite{peng2023trajectory} \raisebox{0ex}{\large{$^{\rm \textcolor{blue}{2023}}$}}    
    &99.8	&193.2	&259.9	&184.3  &24.1	&78.1	&135.8	&79.3  &84.7	&169.7	&251.2	&168.5  &29.9	&78.1	&128.2	&78.7 \\
    
    &JRT~\cite{xu2023joint} \raisebox{0ex}{\large{$^{\rm \textcolor{blue}{2023}}$}} 
    &85.8	&169.3	&233.5	&162.8  &17.3	&68.7	&125.6	&70.5  &\textbf{76.3}	&\textbf{144.4}	&\textbf{213.4}	&\textbf{144.7}  &32.8	&76.2	&118.6	&75.8 \\
    
    &T2P\raisebox{0ex}{\large{$^{\rm \textcolor{blue}{F=1}}$}}~\cite{jeong2024multi} 
\raisebox{0ex}{\large{$^{\rm \textcolor{blue}{2024}}$}}
    &85.8	&178.4	&245.2	&169.8  &17.8	&70.8	&131.7	&73.4  &90.5	&166.8	&235.9	&164.4  &30.3	&79.9	&125.7	&78.6 \\
    
    \rowcolor{gray!20} 
    & \textbf{Ours}   
    &\textbf{72.3}	&\textbf{127.3}	&\textbf{184.8}	&\textbf{128.1}  &15.0	&\textbf{55.9}	&\textbf{103.0}	&\textbf{57.9} &78.4	&151.1	&222.3	&150.6  &\textbf{27.4}	&\textbf{70.4}	&\textbf{114.0}	&\textbf{70.6} \\
    \midrule

    \bottomrule
    \end{tabular}
    }
    \setlength{\abovecaptionskip}{8pt} 
    \setlength{\belowcaptionskip}{-5pt} 
    \caption{Detailed JPE,APE, and FDE results on the CMU-Syn and MuPoTS-3D dataset across different settings, and our model present dominant superiority over most of the compared methods. }
    \label{tab:JPE,APE,FDE on cmu and MuPoTS}
    \end{center}
\end{table*}

\begin{figure*}[!t]
\renewcommand{\thefigure}{B.2} 
\begin{center}
\includegraphics[width=1\linewidth]{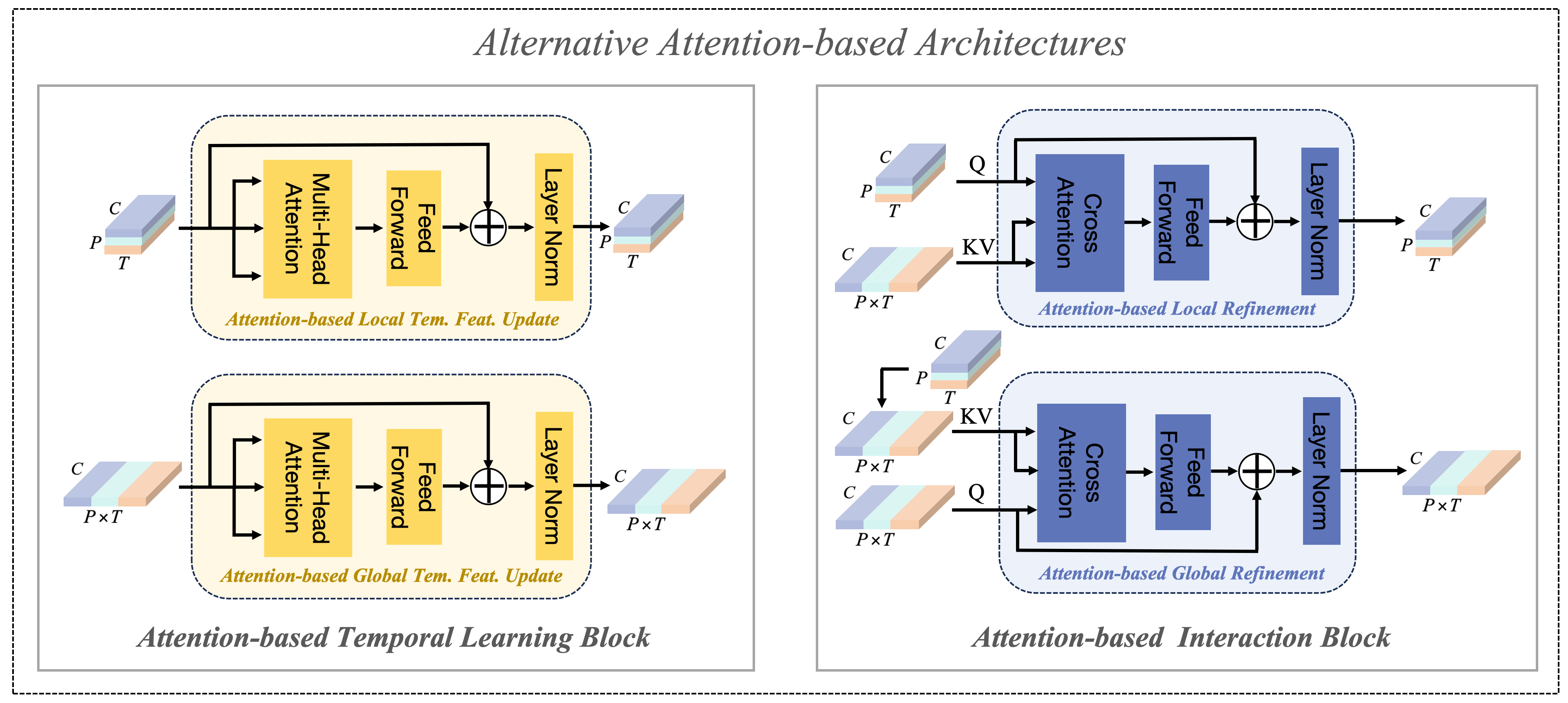}
\end{center}
\setlength{\abovecaptionskip}{-5pt} 
\setlength{\belowcaptionskip}{-10pt} 
\caption{Attention-based architectures used in our ablation study. Our ME block and CI block are replaced with Attention-based Temporal Learning block containing Multi-Head Attention (Self-Attention) module and Attention-based Interaction block containing Cross Attention module,respectively.}
\label{fig:Transformer Architecture}
\end{figure*}

\begin{figure*}[!t]
\renewcommand{\thefigure}{E.1} 
\begin{center}
\includegraphics[width=1\linewidth]{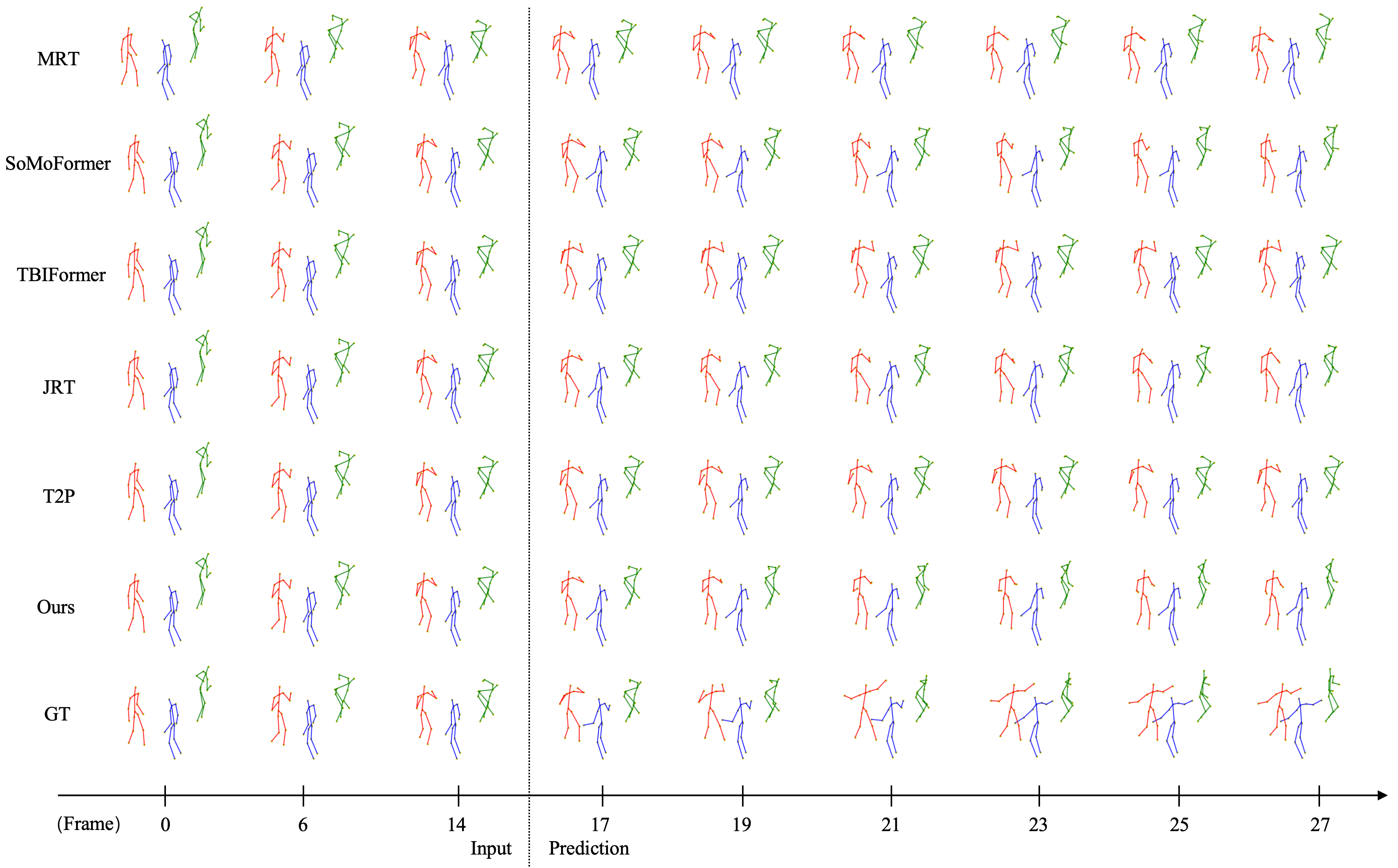}
\end{center}
\setlength{\abovecaptionskip}{-7pt} 
\setlength{\belowcaptionskip}{-10pt} 
\caption{Qualitative results in the \textit{CMU-Syn} (1s/1s) setting. Different colors indicate different individuals. The model predicts 15 frames based on the input of 15 frames.}
\label{fig:mocap15to15}
\end{figure*}

\begin{figure*}[!t]
\renewcommand{\thefigure}{E.2} 
\begin{center}
\includegraphics[width=1\linewidth]{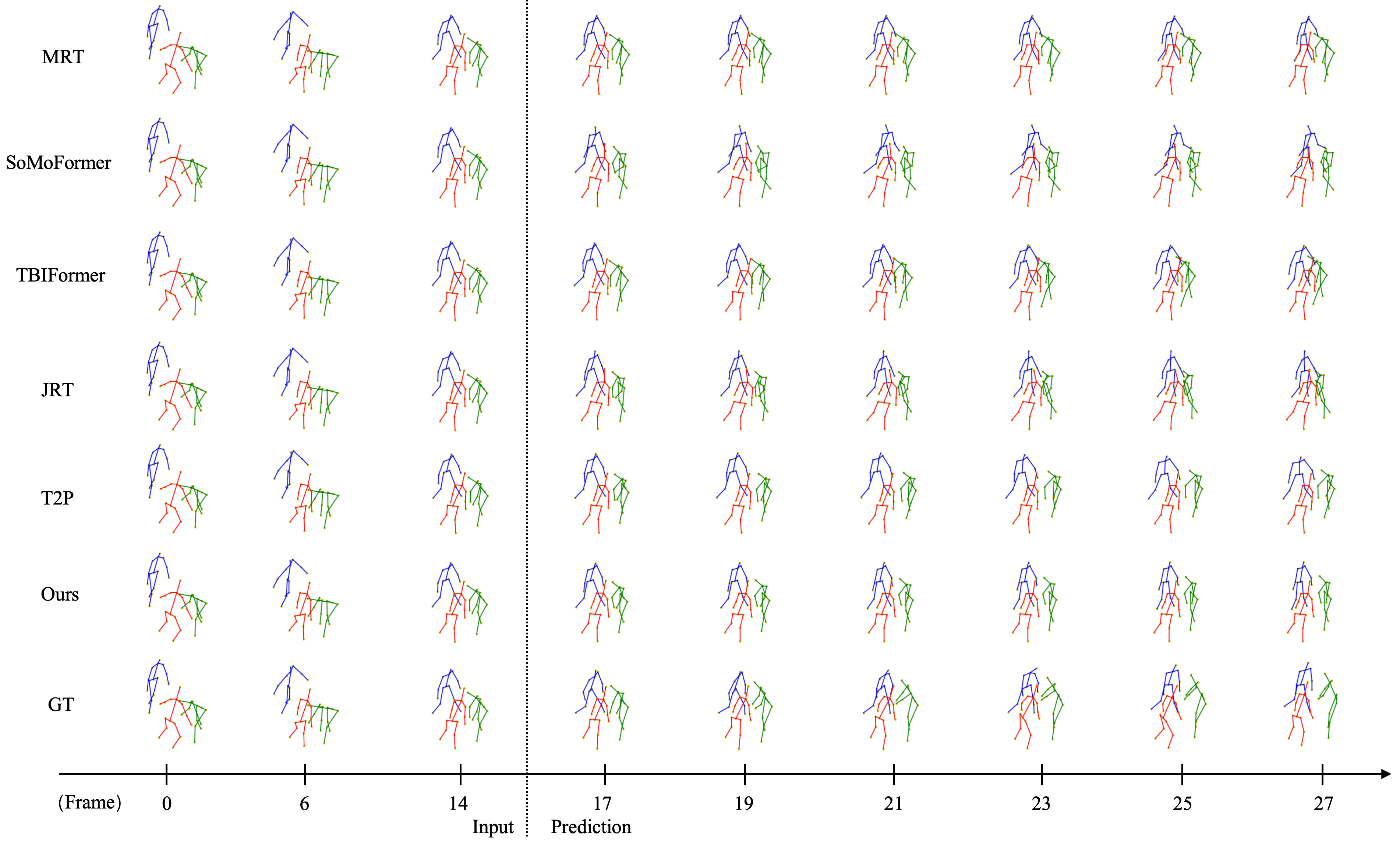}
\end{center}
\setlength{\abovecaptionskip}{-5pt} 
\setlength{\belowcaptionskip}{-20pt} 
\caption{Qualitative results in the \textit{CMU-Syn/MuPoTS} (1s/1s) setting. Different colors indicate different individuals. The model predicts 15 frames based on the input of 15 frames.}
\label{fig:mupots15to15}
\end{figure*}

\clearpage
\clearpage


\end{document}